\documentclass[default,iicol]{sn-jnl}

\jyear{2022}%

\theoremstyle{thmstyleone}%
\newtheorem{theorem}{Theorem}
\theoremstyle{thmstyletwo}%
\theoremstyle{thmstylethree}%
\usepackage[capitalise]{cleveref}
\usepackage{amsmath,amsfonts}
\usepackage{amssymb}
\usepackage{algorithm}
\usepackage{array}
\usepackage{xfrac}
\usepackage{textcomp}
\usepackage{nicefrac}       
\usepackage{microtype}      
\usepackage{booktabs}
\usepackage{mdframed}
\usepackage{accents}
\usepackage{xcolor}
\renewcommand{\vec}[1]{\mathbf{#1}}
\newcommand{\mat}[1]{\mathbf{#1}}
\newcommand{\vz}[0]{\vec{z}}
\newcommand{\vx}[0]{\vec{x}}
\newcommand{\J}[0]{\mat{J}}
\newcommand{\vT}[0]{^{\intercal}}
\newcommand{\dif}[0]{\mathrm{d}}
\definecolor{linecolor}{RGB}{236,222,212} 
\newmdenv[
  leftmargin = 0pt,
  innerleftmargin = 0.5em,
  innertopmargin = 0pt,
  innerbottommargin = 0pt,
  innerrightmargin = 0pt,
  rightmargin = 0pt,
  linewidth = 2pt,
  topline = false,
  rightline = false,
  bottomline = false,
  linecolor=linecolor,
  innertopmargin=0em,
]{leftbar}
\usepackage{mathtools}
\DeclarePairedDelimiter\ceil{\lceil}{\rceil}            
\DeclarePairedDelimiter\floor{\lfloor}{\rfloor}

\newcommand{\KL}[2]{\mathrm{KL}\left( #1 \| #2 \right)} 
\renewcommand{\L}[0]{L(\theta)}                         
\newcommand{\expt}[0]{\mathbb{E}}                       
\newcommand{\lowerOne}[0]{\mathcal{L}(\theta)}
\crefname{equation}{Eq.}{Eqs.}
\crefname{section}{Sec.}{Secs.}
\newcommand{\SV}{\ensuremath{\mathrm{SV}}}
\newcommand{\MI}{\ensuremath{\mathrm{MI}}}
\newcommand{\GP}{\ensuremath{\mathrm{GP}}}
\newcommand{\ZGP}{\ensuremath{\mathrm{0GP}}}
\newcommand{\diff}{\ensuremath{\mathrm{diff}}}
\newcommand{\baseline}{\ensuremath{\mathrm{baseline}}}
\newcommand{\EBM}{\ensuremath{\mathrm{EBM}}}
\newcommand{\Lower}{\ensuremath{\mathrm{lower}}}
\newcommand{\Upper}{\ensuremath{\mathrm{upper}}}
\begin{document}

\title[Article Title]{Exploring bidirectional bounds for minimax-training of Energy-based models}


\author[1]{\fnm{Cong} \sur{Geng}}\email{gengcong@chinamobile.com}
\author*[2]{\fnm{Jia} \sur{Wang}}\email{jiawang@sjtu.edu.cn}
\author[2]{\fnm{Li} \sur{Chen}}\email{hilichen@sjtu.edu.cn}
\author[2]{\fnm{Zhiyong} \sur{Gao}}\email{zhiyong.gao@sjtu.edu.cn}
\author[3]{\fnm{Jes} \sur{Frellsen}}\email{jefr@dtu.dk}
\equalcont{These authors contributed equally to this work.}
\author[3]{\fnm{Søren} \sur{Hauberg}}\email{sohau@dtu.dk}
\equalcont{These authors contributed equally to this work.}

\affil[1]{\orgdiv{China Mobile Research Institute},\orgaddress{ \city{Beijing}, \postcode{100032}, \country{China}}}
\affil[2]{\orgdiv{Institute of Image Communication and Network Engineering}, \orgname{Shanghai Jiao Tong University},\orgaddress{ \city{Shanghai}, \postcode{200240}, \country{China}}}

\affil[3]{\orgdiv{Department of Applied Mathematics and Computer Science}, \orgname{Technical University of Denmark}, \orgaddress{\street{Richard Petersens Plads}, \city{Kgs.\@ Lyngby}, \postcode{2800}, \country{Denmark}}}


\abstract{Energy-based models (EBMs) estimate unnormalized densities in an elegant framework, but they are generally difficult to train. Recent work has linked EBMs to generative adversarial networks, by noting that they can be trained through a minimax game using a variational lower bound. To avoid the instabilities caused by minimizing a lower bound, we propose to instead work with bidirectional bounds, meaning that we maximize a lower bound and minimize an upper bound when training the EBM. We investigate four different bounds on the log-likelihood derived from different perspectives. We derive lower bounds based on the singular values of the generator Jacobian and on mutual information. To upper bound the negative log-likelihood, we consider a gradient penalty-like bound, as well as one based on diffusion processes. In all cases, we provide algorithms for evaluating the bounds. We compare the different bounds to investigate, the pros and cons of the different approaches. Finally, we demonstrate that the use of bidirectional bounds stabilizes EBM training and yields high-quality density estimation and sample generation.}




\maketitle

\section{Introduction}\label{sec1}
Generative models are growing in popularity, especially Generative Adversarial Networks (GANs) \citep{goodfellow2020generative} and Diffusion-based models \citep{ho2020denoising,song2020score}. However, while these can produce high-quality samples, they still have some difficulties with simultaneously explicitly estimating the data density. GANs are able to generate data samples efficiently with the generator, but the discriminator cannot be used to estimate the probability density of the data distribution, which limits their distribution-modeling ability. Diffusion-based models learn the gradients of the log-probability density function~(score function) of data distribution. Then using a Langevin dynamic or ODE solver to generate samples from score functions, which may slow down or complicate sampling. Similar problems appear in density estimation using score functions. 

Energy-Based Models (EBMs) are a class of probabilistic models that parameterize a distribution with an energy function, which provides the possibility for efficient density estimation. Adversarial-based EBMs try to solve the intractable normalization constant in EBMs by introducing a variational distribution so that the generator can be introduced naturally for sampling. It is this class of models we here explore.

EBMs are unnormalized log-densities with a long history in machine learning \cite{Hopfield2554,hinton1983optimal,10.5555/104279.104290}. An EBM defines a probability distribution from the view of a Gibbs density using an energy function $E_{\theta}: \mathcal{X} \rightarrow \mathbb{R}$, which is parameterized by the vector $\theta$:
    \begin{equation}
    p_{\theta}(\vx)=\frac{\exp(-E_{\theta}(\vx))}{Z_{\theta}},
    	\quad
    	Z_{\theta}=\int \exp(-E_{\theta}(\vx)) \dif \vx,
    	\label{eq:ebm}
    \end{equation}
    where $Z_{\theta}$ is the normalization constant (also known as the partition function). This construction can, in principle, model any density with a suitable choice of $E_{\theta}$. We typically learn EBMs using maximum likelihood estimation (MLE), where we seek a value of $\theta$ that minimizes the \emph{negative} data log-likelihood:
    \begin{align}
    \begin{split}
    	\L
    	  :=& -\expt_{\vx \sim p_{\text{data}}(\vx)}\left[\log p_{\theta}(\vx)\right] \\
    	  =& \ \  \expt_{\vx \sim p_{\text{data}}}[E_{\theta}(\vx)] + \log Z_{\theta},
    \end{split}
    \label{eq:neg_loglike}
    \end{align}
    where $p_{\text{data}}$ is the data generating distribution.
    
    The key challenge with EBMs is the lack of a closed-form expression for the normalization constant, $Z_{\theta}$, for non-standard energy functions. This hinders pointwise density evaluation, learning, and sampling. These tasks are, therefore, often performed using approximate algorithms such as Markov Chain Monte Carlo (MCMC) \citep{metropolis1953equation,neal2011mcmc}.
    
\subsection{Minimax games on bounds}
    Recently, a series of papers \cite{kim2016deep, kumar2019maximum, abbasnejad2020gade, che2020your} have linked EBMs and \emph{Wasserstein GANs} (WGANs) \citep{arjovsky2017wasserstein}, such a minimax game forms an approximate learning strategy for EBMs. We here repeat this development following \citet{nomcmc}. First, we develop a lower bound for the log-normalization constant using a proposal (variational) distribution $p_g(\vx)$ followed by Jensen's inequality:
    \begin{align}
        \log Z_{\theta}
        &= \log \expt_{\vx \sim p_g}\! \left[ \frac{\exp(-E_{\theta}(\vx))}{p_g(\vx)} \right] \\
        &\geq \expt_{\vx \sim p_g}\! \left[ \log \frac{\exp(-E_{\theta}(\vx))}{p_g(\vx)} \right] \\
        &= -\expt_{\vx \sim p_g}[E_{\theta}(\vx)] + H[p_g].
        \label{eq:logZbound}
    \end{align}
    Substituting this into \cref{eq:neg_loglike}, we can lower bound the negative log-likelihood as follows:
    \begin{align}
     &\L \geq \lowerOne := \label{eq:lower1}\\
     \begin{split}
     &\expt_{\vx \sim p_{\text{data}}(\vx)}[E_{\theta}(\vx)] - \expt_{\vx \sim p_g(\vx)}[E_{\theta}(\vx)] 
     + H[p_g]. \notag
    \end{split}
    \end{align}
    We emphasize that this bound is tight when $p_g = p_{\theta}$, which is attained when $\lowerOne$ is maximized with respect to $p_g$. When the bound is tight, we can perform maximum likelihood estimation of $\theta$ by minimizing $\lowerOne$. Accordingly, training of an EBM can be seen as a minimax game:
    \begin{align}
      &\min_{E_{\theta}} \max_{p_g} \left\{ \lowerOne \right\} = \label{eq:minimax1} \\
      \begin{split}
      &\min_{E_{\theta}} \max_{p_g} \big\{ \expt_{\vx \sim p_{\text{data}}(\vx)}[E_{\theta}(\vx)] - \expt_{\vx \sim p_g(\vx)}[E_{\theta}(\vx)] \\
      &\hspace{43mm}+ H[p_g] \big\}.\notag
      \end{split}
    \end{align}
    Assuming we can evaluate the entropy of the proposal, we thus have a tractable and elegant way of learning EBMs.
    
\begin{figure}[t]
        \includegraphics[width=\linewidth]{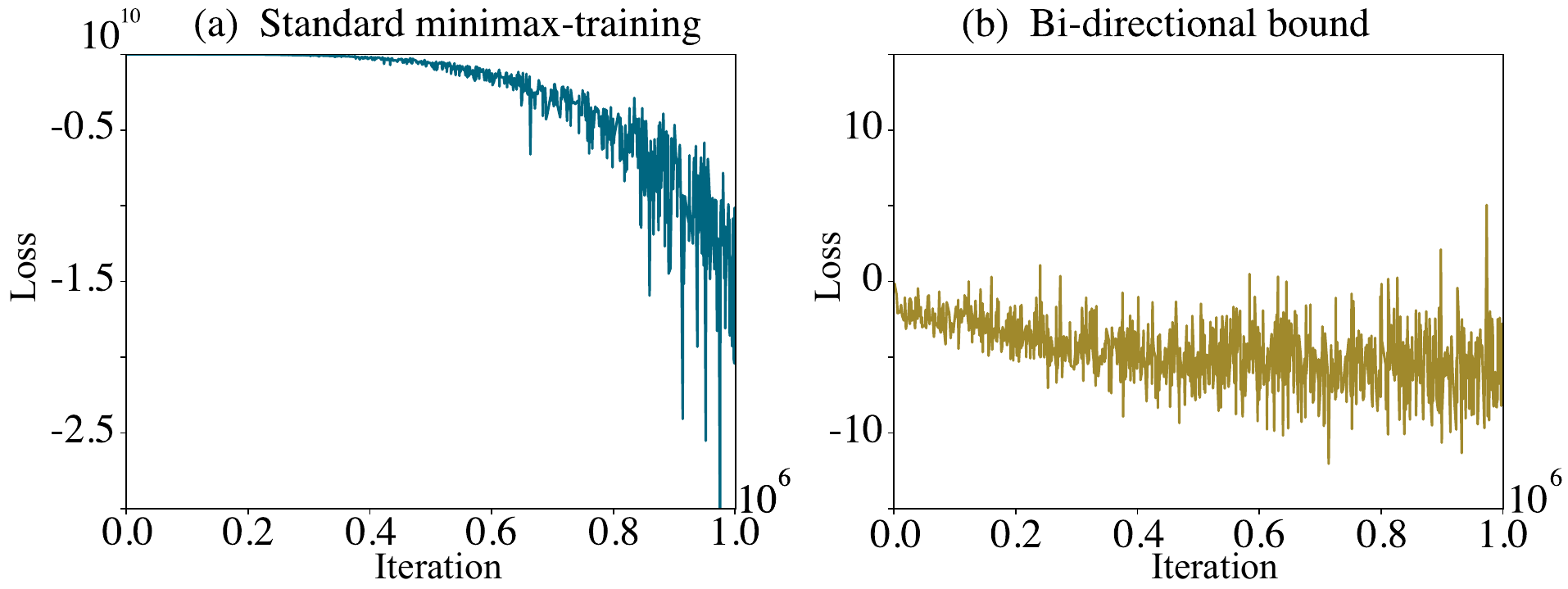}
       
        \caption{Loss function with standard minimax-training (a) or with one of the alternative bounds (b).}
        \label{fig:initial_exp} 
    \end{figure}

\begin{figure}[tb]
            \footnotesize
            \centering
            \renewcommand{\tabcolsep}{1pt} \renewcommand{\arraystretch}{0.1} \begin{tabular}{ccc}
            	\includegraphics[width=0.3\linewidth]{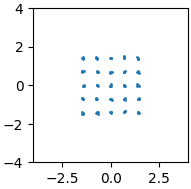} &
            	\includegraphics[width=0.3\linewidth]{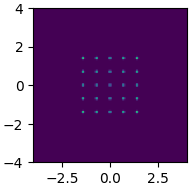} &
            	\includegraphics[width=0.3\linewidth]{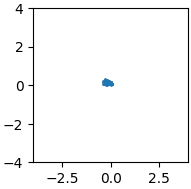} \\
             (a) Data & (b) density  & (c) 1 iter \\
            	\includegraphics[width=0.3\linewidth]{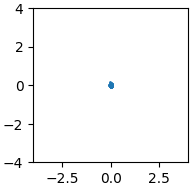}&
            	\includegraphics[width=0.3\linewidth]{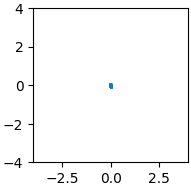} &
            	\includegraphics[width=0.3\linewidth]{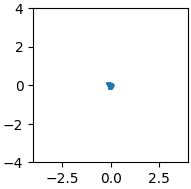} \\
            	
             (d) 5 iters &(e) 10000 iters & (f) 50000 iters 
             \\
            \end{tabular}
            \includegraphics[width=\linewidth]{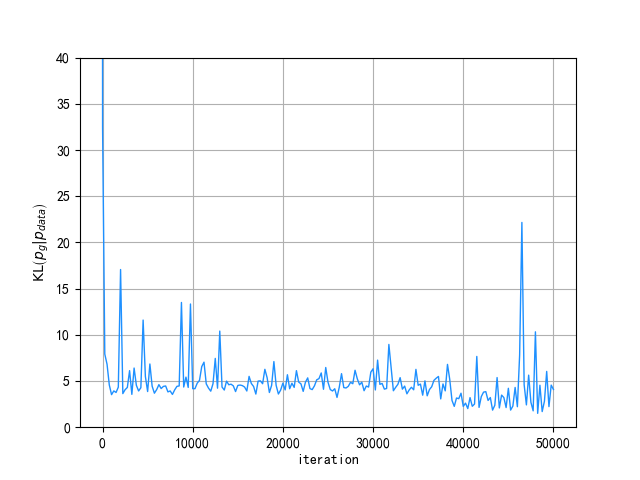}
            \caption{Visualization of generation and plot curve of $\KL{p_g}{p_{data}}$ with different iters when fixing the energy function to be the real log-likelihood.}
            \label{fig:generation_fixed_energy} 
        \end{figure}

    The above analysis, however, assumes that we can tighten the bound by finding a proposal distribution $p_g$ that exactly samples from $p_{\theta}$, i.e.\@ that $p_{\theta}$ is within the variational family. 
    In practice, it is, however, difficult to tighten the lower bound during the training of $p_g$. As illustrated in \Cref{fig:generation_fixed_energy}, if we replace the energy function with the true negative log-likelihood of $p_{data}$, and optimize $p_g$ to tighten the bound, we observe clear mode collapse. We also calculate $\KL{p_g}{p_{data}}$, which is always notably different from zero. This explains why the bound is loose.
    When the bound is \emph{not} tight, as one would generally expect during training, we note that $\lowerOne$ is a \emph{lower bound} on the negative log-likelihood. This is the opposite of conventional \emph{variational inference} \cite{blei2017variational}, where a lower bound on the (non-negative) log-likelihood is maximized. In general, minimizing a lower bound introduces a risk of training towards parameters, where the bound is loosest, instead of where the true objective is near-optimal. In particular, a minimum of the lower bound may be $-\infty$, which is rather unhelpful. To see if this is a problem \emph{in practice}, we fit an EBM to toy data by minimax training using \cref{eq:minimax1}. \Cref{fig:initial_exp} shows that the energy of the training data gradually diverges towards the unhelpful $-\infty$ as training proceeds, suggesting that the potential issues with minimizing a lower bound do appear in practice. In contrast, \cref{fig:initial_exp}b shows the training curve associated with one of our proposed bounds \eqref{eq:upper1}, where it is evident that the divergent training is avoided. 
    
    \textbf{In this paper}, we investigate the issues caused by minimizing a lower bound. We put forward the hypothesis that it is better to minimize an upper bound and maximize a lower bound (\cref{sec:bidi}), rather than playing a minimax game on a single bound. We propose a collection of such \emph{bidirectional bounds} (\cref{sec:lower_bound1,sec:lower_bound2,sec:upper_bound1,sec:upper_bound2}) and compare them empirically (\cref{exp}). We demonstrate that this approach surpasses or matches state-of-the-art on diverse tasks at a negligible computational cost. We position our work to previous ones and touch on historical developments in \cref{sec:related_work}. We note that this paper extends a previous conference paper \cite{geng2021bounds}, where we first proposed the idea of using bidirectional bounds. Here we consider more bounds and perform the associated empirical study.

\section{Minimax-training with bidirectional bounds}\label{sec:bidi}
\begin{figure}[t]
      \begin{center}
        \includegraphics[width=0.5\linewidth]{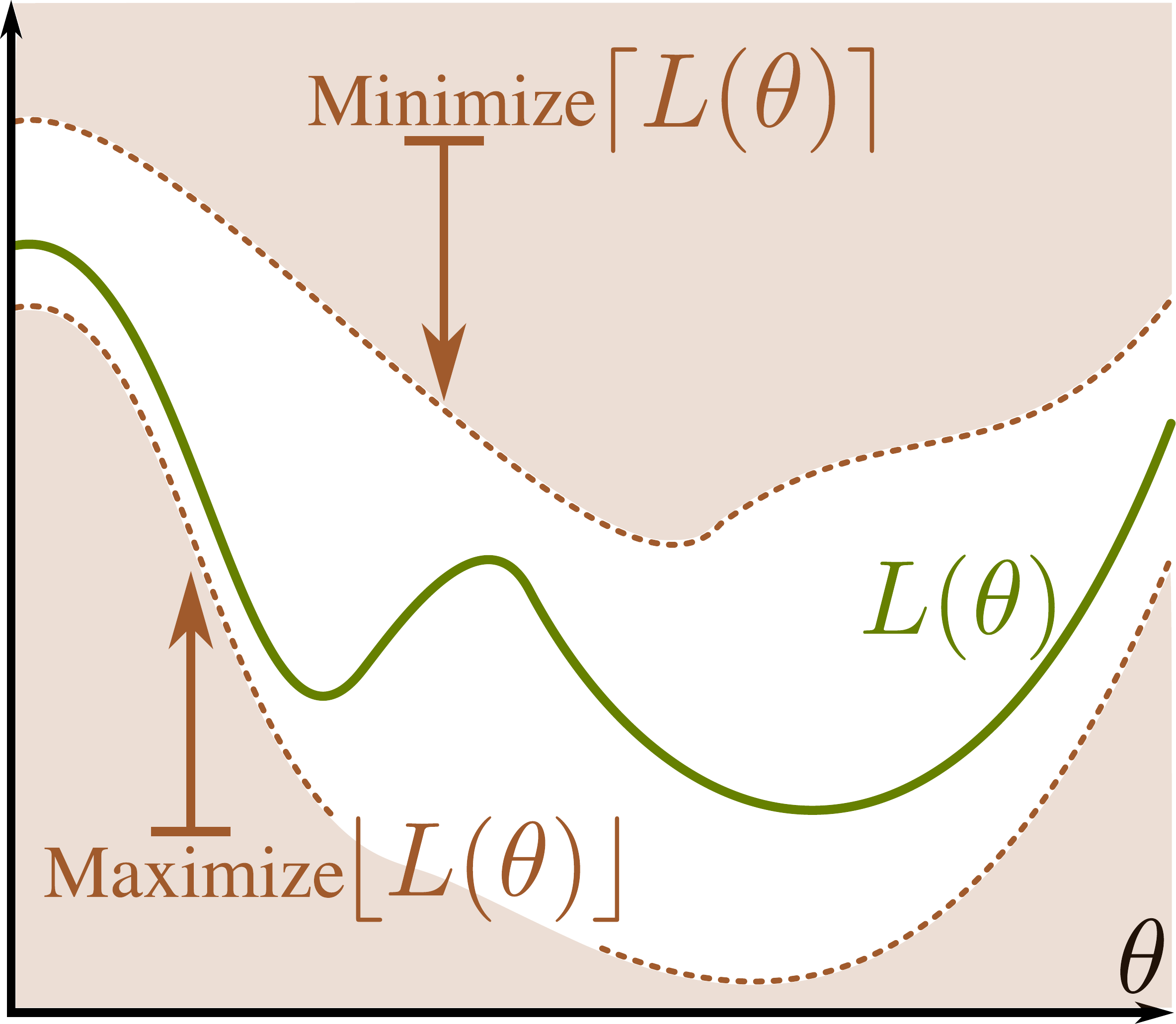}
      \end{center}
      \caption{The bidirectional bounds `sandwich' the negative data log-likelihood, which results in stable training.}
      \label{fig:sandwich}
    \end{figure}

    Our main hypothesis is that instead of solving a minimax game with a lower bounded value function \eqref{eq:lower1}, it is more meaningful to bound $\L$ from both above and below:
    \begin{equation}
      \floor{\L} \leq \L \leq \ceil{\L}.
    \end{equation}
    From this bidirectional bound, we can follow an alternating optimization approach as follows:
    \begin{enumerate}
      \item minimize $\ceil{\L}$ with respect to $E_{\theta}$.
      \item maximize $\floor{\L}$ with respect to $p_g$.
    \end{enumerate}
    This `sandwiching' is illustrated in \cref{fig:sandwich}. We hypothesize that this approach avoids the issues of minimizing a lower bound.
    In this paper, we let $\floor{\L}= \lowerOne$, and propose two approximations of this term. Further, we let $\ceil{\L}=\lowerOne+\ceil{\KL{p_g}{p_{\theta}}}$, where $\ceil{\KL{p_g}{p_{\theta}}}$ denotes an upper bound of $\KL{p_g}{p_{\theta}}$. We use two different methods to find this bound. Below the four different bounds are discussed in detail.

 \subsection{A lower bound from singular values}\label{sec:lower_bound1}
        \cref{eq:lower1} provides a lower bound $\lowerOne$ on $\L$. The first two terms of the bound can be evaluated easily through samples from $p_{\text{data}}$ and $p_g$, respectively. Evaluating the entropy of $p_g$ is not as straightforward. Following common practice in generative adversarial networks (GANs) \citep{goodfellow2020generative}, we define $p_g$ from a base variable $\vz \sim p_0 = \mathcal{N}(\vec{0}, \mat{I})$, which is transformed by a `generator' network $G: \mathbb{R}^d \rightarrow \mathbb{R}^D$, to which we add further noise~\footnote{Refer to \cref{model architecture} for the rationale behind the design of $p_g$.}. Specifically, sampling from $p_g$ becomes
        \begin{align}
          \begin{split}
          \vz &\sim \mathcal{N}(\vec{0}, \mat{I}) \\
          \vx &= G(\vz) + \epsilon, \qquad \epsilon \sim \mathcal{N}(\vec{0}, \sigma_{\text{noise}}^2\mat{I}),
          \end{split}
          \label{eq:generator}
        \end{align}
        where $\sigma_{\text{noise}}^2$ denotes the variance scale of $\epsilon$.
        The generator entropy is, thus, bounded by the entropy of the noise-free sampler, i.e.\@ $H[p_g] \geq H[G(\vz)]$.
        The variable $G(\vz)$ has density over a $d$-dimensional manifold\footnote{Note that \cref{eq:log_p_g} is only the density \emph{on} the spanned manifold, and that the density is zero \emph{off} the manifold.}
        in $\mathbb{R}^D$, which, following standard change-of-variables, is
        \begin{equation}
            \log p (G(\vz)) = \log p_0 (\vz) - \frac{1}{2}\log\det(\J_{\vz}\vT \J_{\vz}),
            \label{eq:log_p_g}
        \end{equation}
        where $\J_{\vz} \in \mathbb{R}^{D \times d}$ is the Jacobian of $G$ at $\vz$.
        This assumes that the Jacobian exists, i.e.\@ that the generator $G$ is differentiable. 
        The entropy of $p_g$ is then bounded from below by\looseness=-1
        \begin{align}
            H[p_g]
              &= -\expt_{\vx \sim p_g}\left[ \log p_g(\vx) \right] \\
              &\geq H[p_0] + \expt_{\vz \sim p_0}\left[ \frac{1}{2}\log\det(\J_{\vz}\vT \J_{\vz}) \right].
        \end{align}
        The entropy of $p_0$ has closed-form expression $H[p_0] = \sfrac{d}{2}(1 + \log(2\pi))$. To avoid the computationally expensive evaluation of the log-determinant-term, we bound this further from below:
        \begin{equation}
            \frac{1}{2} \log\det(\J_{\vz}\vT \J_{\vz})
              = \frac{1}{2} \sum_{i=1}^d \log s_i^2
              \geq d \log s_1,
            \label{eq:entrpy_bound_svd}
        \end{equation}
        where $s_d \geq \ldots \geq s_1$ are the singular values of the Jacobian $\J_{\vz}$. We emphasize that this bound is only tight when all singular values are identical. Combining these developments gives our final lower bound:
        \begin{align}
            \L \geq \floor{\L}
                &= \expt_{\vx \sim p_{\text{data}}(\vx)}[E_{\theta}(\vx)]
	            - \expt_{\vx \sim p_g(\vx)}[E_{\theta}(\vx)] \notag \\
	            &+ H[p_0] + \expt_{\vz \sim p_0}\left[ d \log s_1(\vz) \right]. \hspace{30mm}
	        \label{singular}
        \end{align}

\textbf{Evaluating the lower bound.~~}
The lower bound in \cref{singular} requires access to the smallest singular value of the Jacobian $\J = \partial_{\vz} G(\vz)$. Note that this singular value satisfy $s_1 = \|\J \vec{v}_{\text{min}} \|_2 = \min_{\vec{v} \neq \vec{0}} \frac{\| \J\vec{v}\|_2}{\|\vec{v}\|_2}$. This singular value $\vec{v}_{\text{min}}$ can then be found through optimization, where we choose to use the well-known single-vector \emph{LOPCG algorithm} \cite{knyazev2001toward}. This algorithm iteratively minimizes the generalized Rayleigh quotient,
   \begin{align}
            \rho(\vec{v}) := \frac{\vec{v}\vT \J\vT \J \vec{v}}{\vec{v}\vT \vec{v}},
            \label{eq:rayleigh}
        \end{align}
 which converges to $\vec{v}_{\text{min}}$. The gradient of $\rho(\vec{v})$ is proportional to $r =\J\vT \J \vec{v} - \rho(\vec{v}) \vec{v}$. We use Jacobian-vector products to avoid explicitly forming the Jacobian $\J$ as these can be efficiently using automatic differentiation. To compute $\J\vT \J \vec{v}$, we use a neat trick, which we here specify in \texttt{pytorch}-notation:
        \begin{align}
            \J\vT \J\vec{v}
              = \left((\J\vec{v})\vT \J\right)\vT
              = \nabla_{\vz} \left( (\J\vec{v})\vT\!\!\!\texttt{.detach()} \cdot G(\vz) \right)\vT.
            \label{pro}
        \end{align}
 The optimal step size for this iterative optimization can be found by maximizing the Rayleigh quotient \eqref{eq:rayleigh}. Furthermore, we follow the suggestions of \citet{knyazev2001toward} which improve numerical stability and accelerate convergence. For brevity, we omit these. \looseness=-1
 
 \textbf{Computational cost.~~} The smallest singular value of Jacobian is obtained by an iterative inner loop, which can be costly. For each iteration, we need to compute $\J\vec{v}$ \emph{four} times and $\J\vT \J \vec{v}$ \emph{one} time. Two matrix decompositions are also needed for a $d\times3$ matrix and $D\times3$ matrix respectively.
 Note that we need not differentiate through these operations, which hence are time-consuming but memory efficient. We cannot \emph{priorly} estimate the needed number of iterations of the conjugate gradient~(CG) method\footnote{https://web.cs.ucdavis.edu/~bai/ECS231/Slides/cg\_eig.pdf}, but empirically we have found the estimator to be insensitive to the number iterations. Most likely, this is because we only use the singular value to bound the entropy. In practice, we start from a Hutchinson estimator \citep{hutchinson1989stochastic} which is an upper bound of the entropy, and then use very few iterations to reduce the estimate to a lower bound. We will empirically investigate this in \cref{entropy estimation}.
 
\subsection{A lower bound from mutual information}\label{sec:lower_bound2}
    An alternative estimator of the entropy $H[p_g]$ can be derived through links to mutual information \citep{kumar2019maximum}. Following \citet{kumar2019maximum}, the generator \eqref{eq:generator} entropy can be expressed in terms of the mutual information between $\vx$ and $\vz$ as
    \begin{equation}
      I[\vx, \vz] = H[\vx] - H[\vx \mid \vz].
    \end{equation}
    We note that $H[\vx \mid \vz] = \frac{1}{2}\log(2\pi e\sigma_{\text{noise}}^2)$ is constant. The mutual information $I(\vx,\vz)$ equals the KL divergence between the product of marginals $p_g(\vx)p(\vz)$ and the joint distribution $p(\vx,\vz)$. Rather than estimating this KL divergence, we follow \citet{kumar2019maximum} in using the estimator from \citet{hjelm2018learning}, which computes the Jensen-Shannon divergence between $p_g(\vx)p(\vz)$ and $p(\vx,\vz)$. This choice is based on empirical observations by \citet{kumar2019maximum} that this divergence works better in practice, albeit without implying a tight bound. 
    The estimator then becomes
    \begin{align}
    \begin{split}
      I_{J S D}(\vx, \vz)
        &=\sup _{T \in \mathcal{T}} \mathbb{E}_{p(\vx, \vz)}[-\operatorname{sp}(-T(\vx, \vz))] \\
        &-\mathbb{E}_{p_g(\vx) p(\vz)}[\operatorname{sp}(T(\vx, \vz))],
      \label{eq:lower_mutual}
    \end{split}
    \end{align}
    where $\operatorname{sp}(\cdot)$ is the softplus function, and $T$ is an additional discriminator represented by a neural network.
    As a minor side-remark, we note that \citet{kumar2019maximum} consider a deterministic generator, i.e.\@ $\vx = G(\vz)$, while we add further Gaussian noise. In the case of a deterministic generator, $H(\vx\vert \vz)=-\infty \neq 0$, while the stochastic generator does not have such degeneracy.
    
    \subsection{An upper bound from gradient penalties}\label{sec:upper_bound1}
        As a first upper bound for the log-likelihood, we take inspiration from how WGANs \citep{arjovsky2017wasserstein} are trained as there are strong ties between our lower bound \eqref{eq:lower1} and the WGAN objective,
        \begin{equation}
            L_{\text{WGAN}}
              = \expt_{\vx \sim p_{\text {data }}}[E_{\theta}(\vx)]
              - \expt_{\vx \sim p_g} \left[ E_{\theta}(\vx) \right].
          \label{eq:wgan}
        \end{equation}
        We note that the only difference between \cref{eq:wgan} and \cref{eq:lower1} is the entropy term. We can, thus, expect lessons from WGANs to carry over to the EBM setting. The original WGAN paper \cite{arjovsky2017wasserstein} noted that \emph{gradient clipping} was needed to avoid optimization instabilities. \citet{gulrajani2017improved} proposed to instead add a \emph{gradient penalty} to the loss, such that it becomes
        \begin{align}
        \begin{split}
            L_{\text{WGAN-GP}}
              &= L_{\text{WGAN}} \\
	           &+ \lambda \expt_{\hat{\vx} \sim g(\vx)} \left[(\|\nabla_{\hat{\vx}}E_{\theta}(\hat{\vx})\|_2 - 1)^2\right],
        \end{split}
        \end{align}
        where $\hat{\vx}$ is a uniformly sampled interpolation between real data and generated samples. This is known to be a good way to train WGANs \cite{gulrajani2017improved}. We note that this loss is similar to our lower bound, but that the gradient penalty adds a further positive term. One could, thus, speculate that the result is an upper bound.
        From this intuition, we have proved the following theorem in the conference version of the paper \citep{geng2021bounds}.
        \begin{leftbar}
        \begin{theorem}\label{thm:upper} 
            Suppose $f: \mathcal{X} \rightarrow \mathbb{R}$ is L-Lipschitz continuous, $g(x)$ is a probability density function with finite support, then there exist constants $M, m \geq 0$ and $p \geq 1$ such that:
            \begin{equation}
            \begin{aligned}
                \log \expt_{\vx \sim g(\vx)} \left[f(\vx)\right]-\expt_{\vx \sim g(\vx)} \left[\log f(\vx)\right] \\\leq M (\expt_{\vx \sim g(\vx)} \left[\|\nabla_{\vx}\log f(\vx)\|^p\right]+m)^{\sfrac{1}{p}}.
                 \end{aligned}
                \label{eq:thm_upper}
            \end{equation}
        \end{theorem}
        \end{leftbar}
        Note that the bound in \cref{eq:thm_upper} can be simplified by dropping the power $\sfrac{1}{p}$ if $\expt_{\vx \sim g(\vx)} \left[\|\nabla_{\vx}\log f(\vx)\|^p\right]+m \geq 1$, 
        \begin{equation}
        \begin{aligned}
            \log \left[\expt_{\vx \sim g(\vx)} \left[f(\vx)\right]\right] - \expt_{\vx \sim g(\vx)} \left[\log f(\vx)\right]
            \\\leq
            M\expt_{\vx \sim g(\vx)} \left[\|\nabla_{\vx}\log f(\vx)\|^p\right] + Mm.
            \end{aligned}
            \label{eq:thm_upper_simple}
        \end{equation}
        In our EBM, we set $f(\vx) = \frac{\exp(-E_{\theta}(\vx))}{p_g(\vx)}$ and $g(\vx)=p_g(\vx)$.
        We note that $\log \left[\expt_{\vx \sim g(\vx)} \left[f(\vx)\right]\right] - \expt_{\vx \sim g(\vx)} \left[\log f(\vx)\right]=\KL{p_g}{p_\theta}$, which means that \cref{eq:thm_upper_simple} is an upper bound of $\KL{p_g}{p_\theta}$. We then get an upper bound as $\ceil{\L}=\lowerOne+\ceil{\KL{p_g}{p_\theta}}$.
        
        Empirically, we have found that the bound simplification in \cref{eq:thm_upper_simple} holds throughout most of the training, and therefore use the following (approximate) upper bound:
        \begin{align}
            \ceil{\L}
              &= \lowerOne
               + Mm  \label{eq:upper1}\\
               \begin{split}
              &+ M\expt_{\vx \sim p_g(\vx)} \left[\| \nabla_{\vx} E_{\theta}(\vx) + \nabla_{\vx}\log p_g(\vx)\|^p\right].\notag
        \end{split}
        \end{align}
        %
        For a specific choice of $f(\vx)$, we can view $m$ as a constant, and the term $Mm$ can be ignored during optimization. This developed upper bound has a nice interpretation as the lower bound in \eqref{eq:lower1} plus a gradient penalty. This penalty, however, has a different form than the classic WGAN penalty, which is developed from a regularization perspective. Our proposed upper bound can, thus, provide justification for the commonly used regularization from a maximum likelihood perspective.
        
        \textbf{Bound tightness.~~} When $p_\theta = p_g$, we see that $\left\|\nabla_{\vx}E_{\theta}(\vx)+\nabla_{\vx} \log p_{g}(\vx)\right\|^{p} = 0$. In \cref{thm:upper}, $m$ is a constant related to the Lipschitz constant of $\log f(\vx)$ which satisfy $\left\|\nabla_{\vx} \log f(\vx)\right\|^{p} \leq\left\|\nabla_{\vec{y}} \log f(\vec{y})\right\|^{p}+m$ for all $\vx,\vec{y}$ (see details in the proof). When $p_\theta = p_g$ we furthermore have that $\left\|\nabla_{\vx} \log f(\vx)\right\|^{p} = 0$, such that $m=0$. The upper bound then reduces to $\ceil{\L} = \lowerOne=\L$, which implies that the bound is tight.

        \textbf{Evaluating the upper bound.~~}
        Evaluating \cref{eq:upper1} requires solving two challenges. First, we have to compute $\nabla_\vx\log p_g(\vx)$. Empirically we have found that current methods \cite{shi2018spectral,li2017gradient} are not sufficiently efficient to be practical. Instead, we propose to approximate the term $\expt_{\vx \sim p_g(\vx)} \left[\| \nabla_{\vx} E_{\theta}(\vx) + \nabla_\vx\log p_g(\vx)\|^p\right]$ by further loosening the bound
        \begin{equation}
        \begin{aligned}
            &\| \nabla_{G(\vz)} E_{\theta} (G(\vz)) + \nabla_{G(\vz)}\log p_g(G(\vz))\|^p
             \\ &\leq \frac{\| \nabla_{G(\vz)}E_{\theta} (G(\vz))\J_{\vz}
               + \nabla_{G(\vz)}\log p_g(G(\vz))\J_{\vz}\|^p}{s_1^p}\\
              &\leq \frac{\| \nabla_{G(\vz)} E_{\theta} (G(\vz))\J_{\vz} + \nabla_\vz\log p_g(G(\vz))\|^p}{s_1^p}.
              \label{tranfer}
        \end{aligned}
        \end{equation}
        Here $s_1$ is the smallest singular value of $\J_{\vz}$. 
        In the case $p=2$, we can use Hutchinson's estimator \cite{hutchinson1989stochastic}:
        \begin{equation}
        \begin{aligned}
            &\| \nabla_{\vx} E_{\theta} (\vx)\J_{\vz} + \nabla_\vz\log p_g(G(\vz)) \|_2^2
              \\&= \expt_{\vec{v}}\left[ \left(\nabla_{\vx} E_{\theta} (\vx)\J_{\vz}\vec{v}
              + \nabla_\vz\log p_g(G(\vz))\vec{v}\right)^2 \right],
            \label{eq:gra_esti}
            \end{aligned}
        \end{equation}
        where $\vec{v} \sim \mathcal{N}\left(\vec{0}, \mat{I}_{d}\right)$.
        This can be evaluated efficiently using automatic differentiation.
    
        The second challenge is to compute $\log p_g(\vx)$ which amounts to computing the log-determinant-term $\log\det(\J_{\vz}\vT \J_{\vz})$ of the generator $G(\vz)$ as dictated by \cref{eq:log_p_g}.
        Here, we choose to use our entropy estimator \eqref{eq:entrpy_bound_svd}. 

    \subsection{An upper bound from diffusion}\label{sec:upper_bound2}
        As an alternative upper bound of $\KL{p_g}{p_\theta}$, we take inspiration from \citet{song2021maximum} and introduce a time $t$ to build a diffusion model. If we employ a stochastic differential equation (SDE) to diffuse a distribution $p(\vx)$ towards a noise distribution, we get an SDE of the form:
        \begin{equation}
          \dif \vx =\boldsymbol{f}(\vx, t) \dif t + g(t) \dif \mathbf{w}.
        \end{equation}
        Here we choose the Variance Exploding (VE) SDE \cite{song2020score} for simplicity, which means $\boldsymbol{f}(\vx, t) = 0$, and $g(t)=\sigma_{\min}\left(\frac{\sigma_{\max}}{\sigma_{\min}}\right)^t \sqrt{2 \log \frac{\sigma_{\max}}{\sigma_{\min}}}$. 
        The solution of this SDE is a diffusion process$\{\vx(t)\}_{t \in[0, T]}$, where $[0, T]$ is a fixed time horizon. We let $p_t(\vx)$ denote the marginal distribution of $\{\vx(t)\}$, and $p_{0 t}\left(\vx^{\prime} \mid \vx\right)$
        denote the transition distribution from $\vx(0)$ to $\vx(t)$. Finally, $p_{0 t}\left(\vx^{\prime} \mid \vx\right)$ is a Gaussian distribution where both mean and variance can be computed. If we diffuse $p_g(\vx)$ and $p_\theta(\vx)$ with the same noise distribution, then we can compute an upper bound of the KL divergence \cite{song2021maximum} 
        \begin{align}
          &\mathrm{KL}(p_g(\vx) \| p_\theta(\vx)) \notag \\ 
            &=\KL{p_{gT}(\vx)}{p_{\theta T}(\vx)}
             \!-\! \int_0^T \frac{\partial\KL{p_{gt}(\vx)}{p_{\theta t}(\vx)} }{\partial t}\dif t \notag \\ 
            \begin{split}
            &\leq \ \frac{1}{2} \int_{0}^{T} \mathbb{E}_{p_{gt}(\vx)}
                  \Big[ g^{2}(t)\big\|\nabla_{\vx} \log p_{gt}\left(\vx\right) -\nabla_{\vx}\log p_{\theta t}(\vx)\big\|_{2}^{2}\Big] \dif t \\
            &+\ \KL{p_{gT}(\vx)}{p_{\theta T}(\vx)}.
            \label{upper bound}
            \end{split}
        \end{align}
    If we introduce time $t$ into the energy function, i.e.\@ $p_{\theta t}(\vx)=\frac{\exp(-E_{\theta}(\vx,t))}{Z_{\theta}(t)}$, then $p_{\theta}(\vx)=\frac{\exp(-E_{\theta}(\vx,0))}{Z_{\theta}(0)}$. Combined with score matching~\cite{vincent2011connection}, we immediately obtain:
        \begin{align}
        \begin{split}
          &\KL{p_g(\vx)}{p_\theta(\vx)} \leq \KL{p_{gT}(\vx)}{p_{\theta T}(\vx)} \\
            &+ \frac{1}{2} \int_{0}^{T} \mathbb{E}_{p(\vx) p_{0 t}\left(\vx^{\prime} \mid \vx\right)}\Big[g^{2}(t)\big\|\nabla_{\vx^{\prime}} \log p_{0 t}\left(\vx^{\prime} \mid \vx\right) \\
            &\hspace{40mm} -E_{\theta}^\prime (\vx,t)\big\|_{2}^{2}\Big] \dif t +C,
            \label{improved upper bound}
        \end{split}
        \end{align}
        where $C$ is a constant independent of $\theta$.
        Note that for this to be a valid bound $p_{\theta t}(\vx)$ needs to satisfy Fokker–Planck equation~\cite{appliedSDE}, 
        \begin{equation}
        \frac{\partial p_{t}(\vx)}{\partial t}=\nabla_{\vx} \cdot\left(\frac{1}{2} g^{2}(t) p_{t}(\vx) \nabla_{\vx} \log p_{t}(\vx)\right).
        \end{equation}
        When we directly represent $E_{\theta}(\vx,t)$ with a neural network parameterized by $\theta$ and $t$, it is difficult to ensure that the constraint is satisfied, so we have no strict guarantee that the bound holds.
        Following \citet{song2020improved}, we let $E_{\theta}(\vx,t)=\frac{E_{\theta}(\vx)}{\sigma(t)}$, where $\sigma(t)$ denotes the variance of the transition distribution $p_{0 t}\left(\vx^{\prime} \mid \vx\right)$. \citet{song2020improved} reports that this approximation works well in practice.
        
        Altogether this yields our diffusion-based upper bound
        \begin{align}
        \begin{split}
            &\ceil{\L}=\lowerOne+\ceil{\KL{p_g}{p_\theta}}=\lowerOne\\
            &+\frac{1}{2} \int_{0}^{T} \mathbb{E}_{p(\vx) p_{0 t}\left(\vx^{\prime} \mid \vx\right)}\big[g^{2}(t)\big\|\nabla_{\vx^{\prime}} \log p_{0 t}\left(\vx^{\prime} \mid \vx\right) \\
            &\hspace{40mm} -E_{\theta}^\prime (\vx,t)\big\|_{2}^{2}\big] \dif t \\ 
            &+\KL{p_{gT}(\vx)}{p_{\theta T}(\vx)}+C .
        \end{split}
        \end{align}
        The term $\KL{p_{gT}}{p_{\theta T}}$ is close to 0, if we add enough noise, and we practically ignore the term. The constant $C$ can be ignored during training.
        The integral is computed by calculating the following expectation:
         \begin{align}
            &\frac{1}{T} \int_{0}^{T} p(t)\mathbb{E}_{p(\vx) p_{0 t}\left(\vx^{\prime} \mid \vx\right)}\big[g^{2}(t)\big\|\nabla_{\vx^{\prime}} \log p_{0 t}\left(\vx^{\prime} \mid \vx\right) \notag \\
            &\hspace{35mm} -E_{\theta}^\prime (\vx,t)\big\|_{2}^{2}\big] \dif t,
            \label{compute_diffusion}
         \end{align}
       where $p(t)$ is uniform over $[0,T]$. Computationally, we sample $t \sim p(t)$ to evaluate \cref{compute_diffusion} using Monte Carlo. Since $\log p_{0 t}\left(\vx^{\prime} \mid \vx\right)$ is Gaussian, the computation of \cref{compute_diffusion} is efficient \citep{song2021maximum}.
       
        \textbf{Bound tightness.~~} If $p_g$ and $p_\theta$ have finite second moments and continuous second-order derivatives, then our bound is tight \citep{song2021maximum}. Furthermore, if $p_g = p_\theta$, then \cref{upper bound} is easily seen to 0 such that $\ceil{\L} = \lowerOne= \L$ and our bound becomes tight.\looseness=-1

    \subsection{Relevance for Cooperative Networks}
         Cooperative Networks~(CoopNets) employ cooperative training to train both the generator and energy function. To optimize the energy function, CoopNets initializes samples from $p_{\theta}$ using the generator's output, which significantly accelerates MCMC sampling.
         As mentioned by \citep{xie2018cooperative2}, let $\mathcal{M}_{\theta}^{(k)}$ denote $k$ steps MCMC transition kernel, then $\mathcal{M}_{\theta}^{(k)} p_g$ is the approximated distribution of $p_{\theta}$ after MCMC refinement. Therefore, the energy training objective is
         \begin{equation}
            \mathrm{KL}\left(p_{\text {data }} \| p_\theta\right)-\mathrm{KL}\left(\mathcal{M}_{\theta}^{(k)} p_g \| p_\theta\right). \label{coop_energy}
        \end{equation}
        We see that CoopNets still minimize a lower bound, even though the difference between this bound and $\mathrm{KL}\left(p_{\text {data }} \| p_\theta\right)$ is tighter than that between $\lowerOne$ and $\L$. 
        When optimizing the generator, we minimize $\mathrm{KL}(p_g \| p_\theta)$, while CoopNets minimize $\mathrm{KL}(\mathcal{M}_{\theta}^{(k)} p_g  \| p_g)$. The CoopNet objective may have some advantages regarding mode collapse, but it needs MCMC sampling from the posterior $p_g(z\mid x)$, which can be costly and will depend on the settings of MCMC sampling.

 \section{Related work}\label{sec:related_work}
    \subsection{Deep generative EBMs}
    In machine learning, interests in EBMs can be dated back to Hopfield networks \cite{Hopfield2554}, Boltzmann machines \cite{hinton1983optimal,hinton1985learning, osogami2017boltzmann} and restricted Boltzmann machines \cite{10.5555/104279.104290,hinton2002training}, see e.g.\@ reviews in \cite{lecun2006tutorial} and \cite{scellier2020deep}. It is difficult to learn and evaluate these models because of the expensive computation of the normalization constant. Some MLE-based learning algorithms use expensive MCMC-based methods to estimate the gradient or the normalization constant, see e.g.\@ \citep{salakhutdinovicml08a,grosse2013annealing,pmlr-v38-burda15,pmlr-v51-frellsen16}. Moreover, some non-deep EBMs also employed MCMC-based learning on texture modeling \cite{zhu1998filters,zhu1998grade} and natural image patterns \cite{xie2015learning,xie2016inducing}. To avoid the costly MCMC approximation, \citet{nijkamp2019learning} proposed a short-run MCMC procedure for EBMs and later described an interesting way to check whether EBM is well-learned with MCMC \citep{nijkamp2020anatomy}.
    \citet{hinton2002training} proposed a $k$-step Contrastive Divergence (CD-$k$) method to approximate the negative phase log-likelihood gradient. \citet{hyvarinen2005estimation} proposed an alternative method that used score matching to train non-normalized graphical models. Subsequently, deep EBMs such as deep belief networks \citep{hinton2006fast} and deep Boltzmann machines \citep{pmlr-v5-salakhutdinov09a}, have been proposed.
        
    Recently, deep generative EBMs have grown in popularity, especially for image generation. \citet{nomcmc} outlined three classes of learning-based methods for their current developments and obvious weaknesses, which we summarize here: (1)~MLE methods with MCMC sampling are slow while (2)~score-matching-based methods are comparable faster but do not work with discontinuous nonlinearities, both of them are unstable. (3)~Noise-contrastive estimation \citep{gutmann2010noise} relies on density ratio estimation to learn energy functions but typically does not scale well with the data dimensionality.
        
    Our proposed method belongs to a fourth class of methods that uses a simultaneously learned generator or variational distribution to avoid costly MCMC sampling.
    For this class, \citet{kim2016deep} also proposed to use a generator function and update the generator using the same lower bound as us, but a different entropy approximation. Furthermore, without an extra regularizer like our upper bound when optimizing the energy function, their energy function needs to be explicitly designed to prevent it from growing to infinity which limits its potential.

\subsection{Adversarial-based EBMs}  
    In the past years, generative models based on adversarial training have made significant gains through variations of the GAN \citep{goodfellow2020generative}. Unlike EBMs, these GAN-style models are not explicit likelihood models. It is worth noting that the only difference between the WGAN loss \citep{arjovsky2017wasserstein} and the lower bound in \cref{eq:lower1} is the entropy term. We can, thus, expect lessons from the WGAN to apply to EBMs as well. \citet{xie2017synthesizing,xie2018learning} and \citet{wu2018sparse} were among the first to employ adversarial training of EBMs. \citet{zhai2016generative} proposed to use the same lower bound as us with a minimax game to jointly optimize the energy function and generator. However, a specifically designed bounded multi-modal energy function is needed for their methods which limits its potential. Furthermore, their methods provide no theoretical guarantees on the approximation quality of the generator entropy, and their formulation does not include an upper bound on the regularization of the energy function.

    \citet{dai2017calibrating} also proposed to jointly learn the energy function and a generator using an adversarial learning framework. Two approaches were employed to maximize the generator entropy: One is to maximize the entropy by minimizing the conditional entropy using a variational upper bound, and another is to assume the data is isotropic Gaussian, which is not suitable for high-dimensional data.
    \citet{kumar2019maximum} also consider adversarial learning with an estimator of the entropy of the generator that is based on mutual information. Our lower bound from mutual information \eqref{eq:lower_mutual} is nearly identical to the estimator from \citet{kumar2019maximum}, with the only difference being that our generator is stochastic.
    \citet{abbasnejad2019generative} approximated the entropy using the generator function’s Jacobian log-determinant. However, their method does not scale to high-dimensional data.
    \citet{han2019divergence} proposed a divergence triangle loss with an adversarial learning strategy, whose training mechanism is quite different from ours. An extra encoder is also needed for learning the generator.

    A \emph{cooperative learning} between the energy function and a generator was proposed by \citet{xie2018cooperative, xie2018cooperative2, xie2021cooperative, xie2021learning}. 
    However, this type of approach relies on Langevin dynamics or MCMC to draw samples from the EBM, which can be expensive and difficult to tune.
    \citet{nomcmc} proposed VERA to avoid the use of MCMC which is similar to our work. But unlike us, VERA plays a minimax game both on a lower bound and approximates the gradient of the entropy term using variational inference. Furthermore, VERA is memory-consuming and uses a zero-centered gradient penalty as a regularizer for the energy function, which is a heuristic without an explicit theoretical guarantee. 
    
\section{Experiments}\label{exp}
    To demonstrate the efficiency of training EBMs with bidirectional bounds, we compare against a selection of methods for different model classes. To compare against GANs, we use \emph{deep convolutional GANs (DCGANs)} by \citet{radford2015unsupervised}, \emph{spectrally normalized GANS (SNGANs)} by \citet{miyato2018spectral} and the \emph{WGAN with zero-centered gradient penalty (WGAN-0GP)} by \citet{thanh2019improving}. To compare against other EBMs, we use the \emph{maximum entropy generators (MEG)} by \citet{kumar2019maximum} and the \emph{variational entropy regularized approximate maximum likelihood (VERA)} estimator \citep{nomcmc}. 
    To compare against CoopNets, we use two similar methods~\citep{xie2018cooperative,xie2021cooperative} and \emph{EBM-VAE}  \citep{xie2021learning}. We also consider \emph{DDPM} \citep{ho2020denoising} and \emph{NCSN} \citep{song2019generative} to represent diffusion-based models and score-matching models, respectively.
    
    For bidirectional bounds, we consider all four combinations of our proposed bounds to determine which works best in practice. To differentiate the combinations, we denote
    \begin{itemize}
      \item $\SV$ as the lower bound derived from singular values \eqref{singular},
      \item $\MI$ as the mutual information lower bound \eqref{eq:lower_mutual},
      \item $\GP$ as the gradient penalties upper bound \eqref{eq:upper1}, and
      \item $\diff$ as the score-based diffusion upper bound \eqref{improved upper bound}.
    \end{itemize}
    We then denote combinations using $\EBM_{\Lower+\Upper}$, e.g.\@ $\EBM_{\SV+\GP}$. It is worth noting that both the lower bound based on singular values and the upper bound based on gradient penalties require computing $H(p_g)$. Since this is computationally expensive, we generally recommend using these bounds jointly in order to reduce the total computational cost. To get further insight into the influence of using an upper bound, we introduce another baseline in which we use a zero-centered gradient penalty instead of an actual upper bound. This is similar in spirit to our $\GP$ upper bound. We denote this model $\EBM_{\ZGP}$.
    \subsection{Training details}\label{training details}
        On the small toy data, we realize all models using multi-layer perceptrons (MLPs), while for challenging datasets, we use the same convolutional architecture as \texttt{StudioGAN} \citep{kang2020ContraGAN}.
        The discriminator network $T$ in mutual information-based lower bound \eqref{eq:lower_mutual} follows the design of \citet{kumar2019maximum}.
        We conduct all experiments on a single 12GB NVIDIA Titan GPU using \texttt{pytorch}  \citep{paszke2017automatic}. We found that training is improved when we use a positive margin for our energy function of gradient penalties upper bound to balance the terms. Specifically, 
        \begin{equation}
        \begin{aligned}
            \ceil{\L}&= \floor{\L} \\
            &+ \big[ M\expt_{\vx \sim p_g(\vx)} \left[\| \nabla_{\vx} E_{\theta} (\vx)
                     + \nabla_{\vx}\log p_g(\vx)\|^p\right]- \zeta \big]_+, 
              \label{training upper bound}
        \end{aligned}
        \end{equation}
        where $[\cdot]_+ = \max(0, \cdot)$ is the usual hinge. In most experiments, we set $\zeta = 1$ as this lets us use the simplified bound in \cref{eq:upper1}. Generally, we recommend starting with $\zeta = 0$ and successively considering larger values until training stabilizes. For $\EBM_{\SV+\GP}$, like~\cite{nomcmc,dieng2019prescribed}, we also find it helpful to add an entropy regularizer weight $\lambda$, which is consistent with theory when set to 1. In larger-scale experiments, we reduced the weight of the entropy regularizer to improve performance. For $\EBM_{\MI+\diff}$, we choose the time horizon $[0,T]$ for diffusion process as [0,1]. $\sigma_{min}=0.01$ and $\sigma_{max}$ is adjusted to different experiments. But we set $\sigma_{max}=0.1$ for most experiments.
        
         \begin{figure*}[tb]
            \footnotesize
            \centering
            \renewcommand{\tabcolsep}{1pt} \renewcommand{\arraystretch}{0.1} \begin{tabular}{ccccccc}
            	\includegraphics[width=0.14\linewidth]{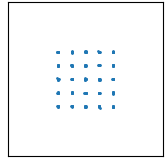} &
            	\includegraphics[width=0.14\linewidth]{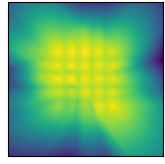} &
            	\includegraphics[width=0.14\linewidth]{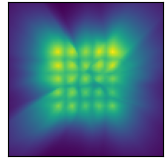} &
            	\includegraphics[width=0.14\linewidth]{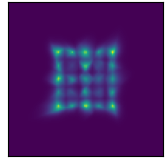}&
            	\includegraphics[width=0.14\linewidth]{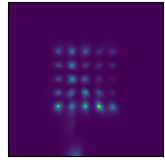} &
            	\includegraphics[width=0.14\linewidth]{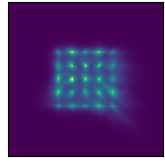} &
            	\includegraphics[width=0.14\linewidth]{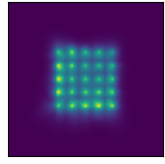}
            	\\
            	\includegraphics[width=0.14\linewidth]{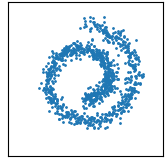} &
            	\includegraphics[width=0.14\linewidth]{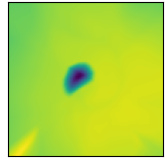} &
            	\includegraphics[width=0.14\linewidth]{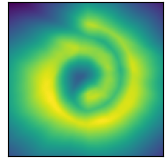} &
            	\includegraphics[width=0.14\linewidth]{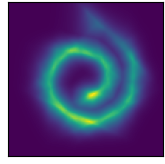}&
            	\includegraphics[width=0.14\linewidth]{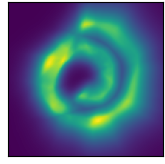} &
            	\includegraphics[width=0.14\linewidth]{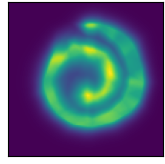} &
            	\includegraphics[width=0.14\linewidth]{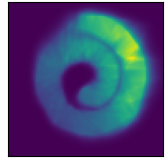} 
            	\\
            	(a) Data & (b) WGAN-0GP & (c) MEG & (d) VERA &(e) $\EBM_{\ZGP}$ & (f) $\EBM_{\SV+\GP}$ & (g) $\EBM_{\MI+\diff}$
            \end{tabular}
            \caption{Density estimation on the 25-Gaussians and swiss-roll datasets.}
            \label{density} 
        \end{figure*}

        \begin{figure*}[h]
            \footnotesize
            \centering
            \renewcommand{\tabcolsep}{1pt} \renewcommand{\arraystretch}{0.1} \begin{tabular}{ccc}
            \includegraphics[width=0.33\linewidth]{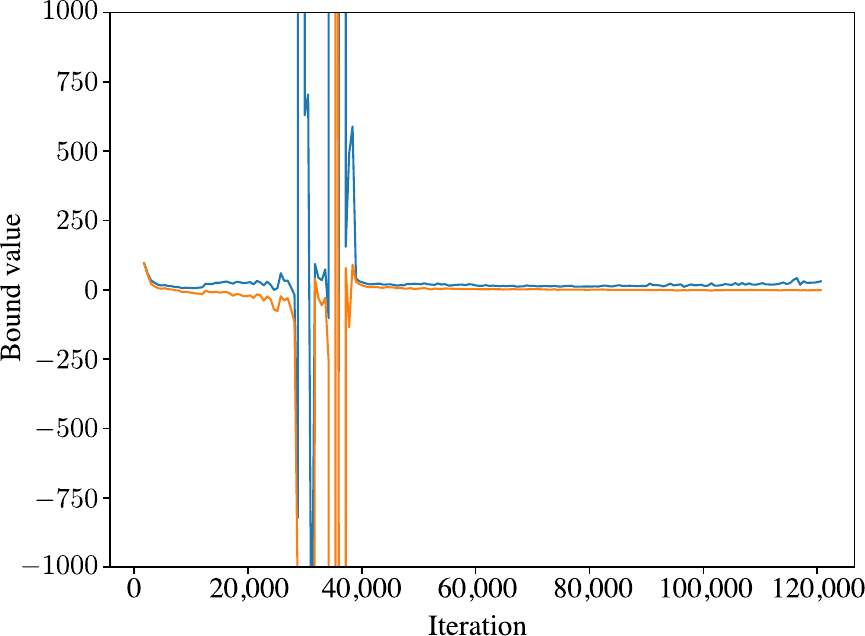}&
            \includegraphics[width=0.33\linewidth]{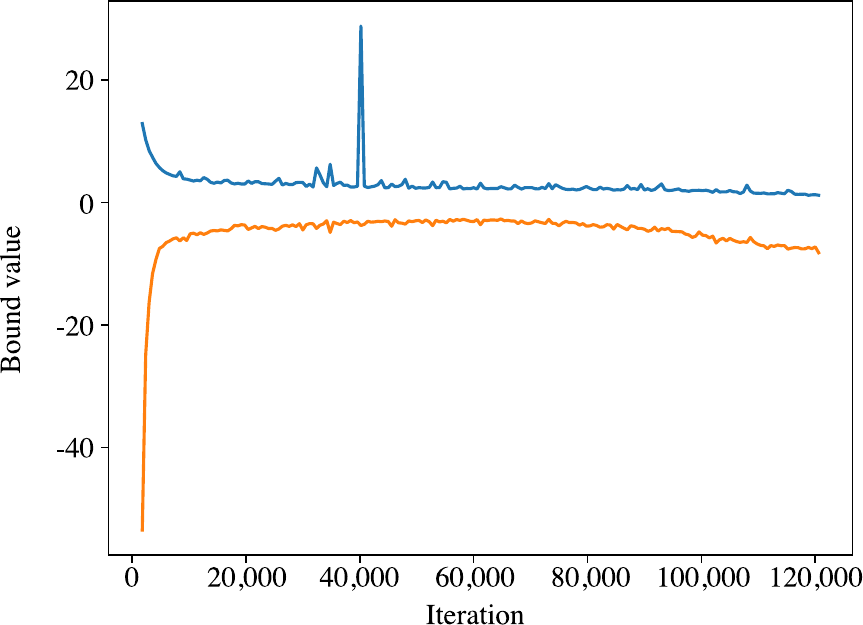}&
            \includegraphics[width=0.33\linewidth]{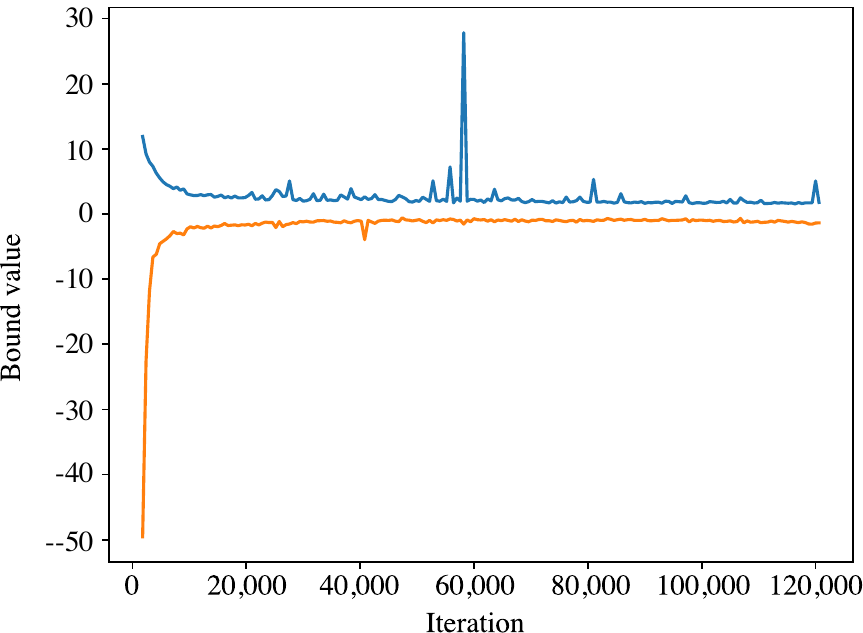} 
            	\vspace{1mm}\\
            (a) $\EBM_{\baseline}$ &	(b) $\EBM_{\SV+\GP}$ & (c) $\EBM_{\MI+\GP}$
            \vspace{1mm}\\
             \end{tabular}
            \begin{tabular}{cc}
             \includegraphics[width=0.33\linewidth]{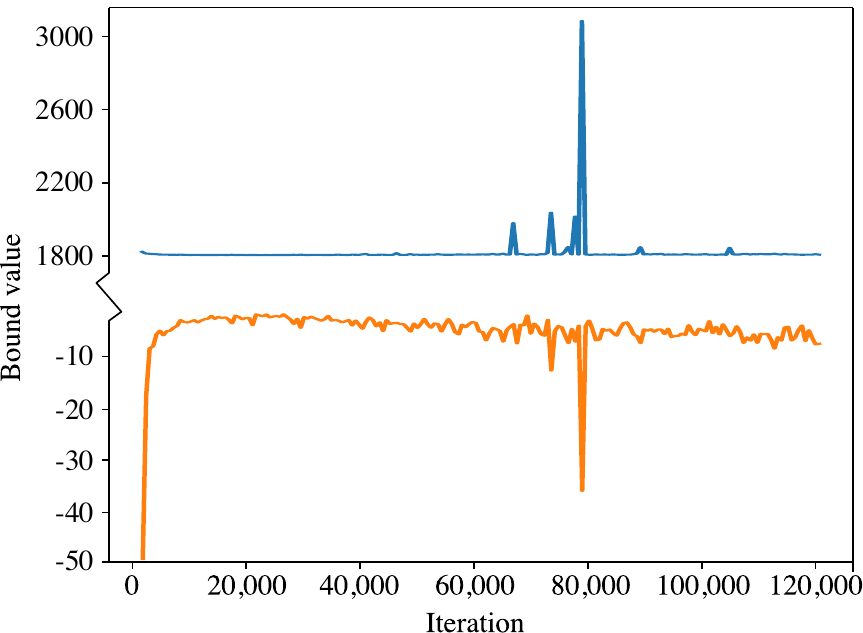}
                \hspace{3mm} & \hspace{3mm}
            	\includegraphics[width=0.33\linewidth]{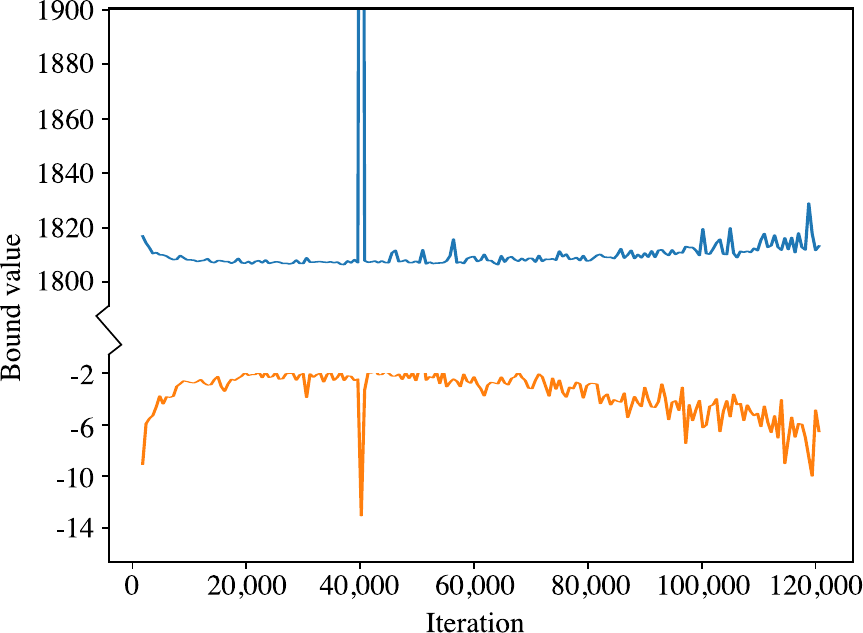} 
            \vspace{1mm}\\
            	(c) $\EBM_{\SV+\diff}$ & (d) $\EBM_{\MI+\diff}$ \\
            \end{tabular}
       \begin{tabular}{ccccc}
       \includegraphics[width=0.2\linewidth]{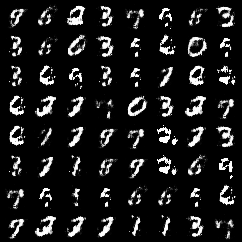} &
            	\includegraphics[width=0.2\linewidth]{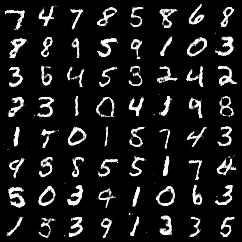} &
            	\includegraphics[width=0.2\linewidth]{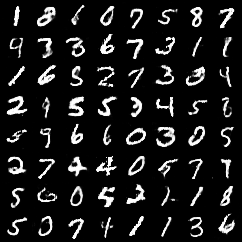} &
             \includegraphics[width=0.2\linewidth]{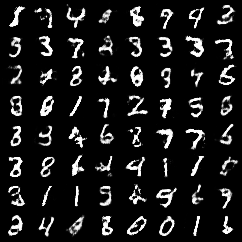} &
            	\includegraphics[width=0.2\linewidth]{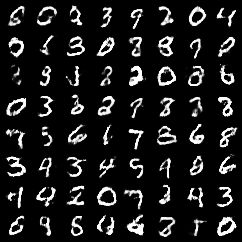} \vspace{1mm}\\
            	(a) $\EBM_{\baseline}$ & (b) $\EBM_{\SV+\GP}$ & (c) $\EBM_{\MI+\GP}$ & (d) $\EBM_{\SV+\diff}$ & (e) $\EBM_{\MI+\diff}$ 
            \end{tabular}
            \caption{Bidirectional bounds estimation for each combination of different bounds and their corresponding generation results.
              }
            \label{bounds} 
        \end{figure*}

\subsection{Toy data}
        We first experiment with the 25-Gaussians and swiss-roll datasets, and \cref{density} shows estimated densities using our methods, WGAN-0GP, and two baselines. We observe that WGAN's discriminator does not provide a useful density estimate. This is unsurprising as WGAN is not optimized to provide such. MEG and $\EBM_{\ZGP}$ estimate the density inaccurately in some boundary and peripheral areas. This may be caused by insufficient or excessive optimization of entropy as the zero-centered gradient penalty is not derived from the maximum likelihood principle. Both our $\EBM_{\SV+\GP}$ and VERA successfully learn a sharp distribution, but our method is more stable in some inner regions of the 25-Gaussians and boundary regions of the swiss-roll. $\EBM_{\MI+\diff}$ generates clearer details on 25 Gaussian but is less sharp on swiss-roll.
        
\subsection{MNIST}
        \noindent\textbf{Mode counting.~~}
        Mode collapse is a common issue in generative models, where the generator $G$ maps all latent inputs to a reduced number of observation points. GANs, in particular, suffer from this issue. EBMs train the generator to have maximal entropy, and thus they often do well at capturing all modes of the data distribution. We follow the test procedure of \citet{kumar2019maximum} to empirically verify that this is also the case for our EBMs. We train all models on the StackedMNIST dataset, which is a synthetically created dataset that stacks MNIST on different channels. The ground truth total number of modes is $1{,}000$, which can empirically be counted using a pre-trained MNIST classifier. We further calculate the KL divergence between the generated mode distribution and the data distribution, and present both results in \cref{mode_counting}. As one would expect, GAN-based methods generally suffer from mode collapse with the exception of WGAN-0GP. The EBMs generally capture (nearly) all modes. Our $\EBM_{\ZGP}$ finds all modes and achieves the smallest KL divergence. It is worth emphasizing that WGAN-0GP obtains comparable results, which aligns with observations by \citet{nomcmc}. We also observe that our $\EBM_{\SV+\GP}$ is sensitive to the choice of network structure, as it only gives good performance with a fully connected network, while it suffers from mode-collapse when using a CNN network. $\EBM_{\MI+\diff}$, on the other hand, gives consistently good results across network architectures.
         \begin{table}[tb]
         \centering
            \caption{Number of captured modes and KL divergence between the real and generated distributions.}
            \label{mode_counting}
            \begin{tabular}{cccccc}
                \toprule
                \multicolumn{2}{c}{Model} & \multicolumn{2}{c}{Modes$\uparrow$} & \multicolumn{2}{c}{KL$\downarrow$} \\ \midrule
                \multicolumn{2}{c}{DCGAN} & \multicolumn{2}{c}{392 $\pm$ 7.4} & \multicolumn{2}{c}{8.012 $\pm$ 0.056} \\
                \multicolumn{2}{c}{SNGAN} & \multicolumn{2}{c}{441 $\pm$ 39.0} & \multicolumn{2}{c}{2.755 $\pm$ 0.033 } \\
                \multicolumn{2}{c}{WGAN-0GP} & \multicolumn{2}{c}{\textbf{1000 $\pm$ 0.0}} & \multicolumn{2}{c}{0.048 $\pm$ 0.003} \\
                \multicolumn{2}{c}{MEG} & \multicolumn{2}{c}{\textbf{1000 $\pm$ 0.0}} & \multicolumn{2}{c}{0.042 $\pm$ 0.004} \\
                \multicolumn{2}{c}{VERA} & \multicolumn{2}{c}{989 $\pm$ 9.0} & \multicolumn{2}{c}{0.152 $\pm$ 0.037} \\
                \multicolumn{2}{c}{$\EBM_{\ZGP}$} & \multicolumn{2}{c}{\textbf{1000 $\pm$ 0.0}} & \multicolumn{2}{c}{\textbf{0.039 $\pm$ 0.003}}\\
                \multicolumn{2}{c}{$\EBM_{\SV+\GP}$~(FC)} & \multicolumn{2}{c}{\textbf{1000 $\pm$ 0.0}} & \multicolumn{2}{c}{0.045 $\pm$ 0.003} \\ 
                \multicolumn{2}{c}{$\EBM_{\SV+\GP}$~(conv)} & \multicolumn{2}{c}{-} & \multicolumn{2}{c}{-} \\
                \multicolumn{2}{c}{$\EBM_{\MI+\diff}$~(FC)} & \multicolumn{2}{c}{988$\pm$2.36} & \multicolumn{2}{c}{0.401$\pm$0.040} \\
                \multicolumn{2}{c}{$\EBM_{\MI+\diff}$~(conv)} & \multicolumn{2}{c}{\textbf{1000 $\pm$ 0.0}} & \multicolumn{2}{c}{0.045 $\pm$ 0.003} \\
                \bottomrule
            \end{tabular}
        \end{table}
        \begin{table}[tb]
            \centering
            \caption{MNIST test log-likelihood on energy function with upper and lower bounds combinations}
            \label{log-likelihood}
            \begin{tabular}{ccc}
                \toprule
                \multicolumn{1}{l}{} & \multicolumn{1}{c}{$\MI$} & $\SV$ \\ \midrule
                $\ZGP$ & $-1819\pm 35.8$ & $-2167\pm 507.2$ \\
                $\GP$ & $\mathbf{-1782\pm 132.6}$ & $-1634\pm76.0$ \\
                $\diff$ & $-1798\pm 281.8$ & $\mathbf{-1561\pm 297.5}$ \\ \bottomrule
            \end{tabular}
        \end{table}

\begin{figure}[tb]
            \centering
        	\includegraphics[width=0.8\linewidth]{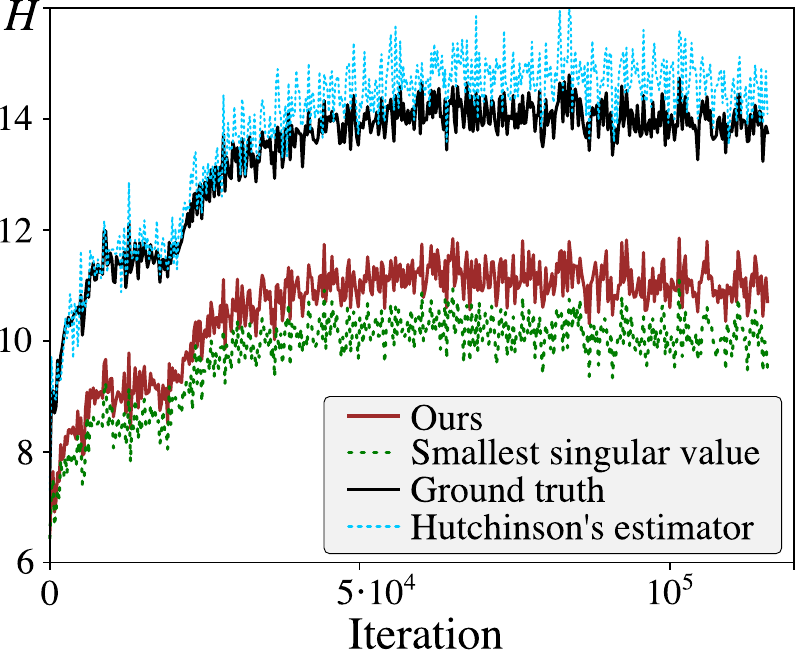}
        	\caption{Entropy estimators with Hutchinson's estimator, true smallest singular value based estimator, and our approximation in singular value lower bound.}
        	\label{entropy estimation}
        \end{figure}

 \textbf{Entropy estimation.~~}
        Next, we evaluate how well our entropy estimator based on singular values performs when measured on MNIST (\cref{entropy estimation}). The lower bound derived from mutual information uses the Jensen-Shannon divergence rather than the KL divergence, which hinders direct entropy estimation. Hence we do not include this in the comparison. We here consistently use networks with fully-connected layers as we can then easily derive the Jacobian in closed form, which provides us with the ground truth entropy of the generator. We compare our estimator (red) with the ground truth (black) and the variant of Hutchinson's estimator (blue) proposed by \citet{kumar2020regularized}. Finally, to investigate the choice of only taking a few steps in our calculation of the singular values, we also compare with a high-precision estimator of the smallest singular value (green).
        We see that the variant of Hutchinson's estimator is quite close to the ground truth, but that it consistently upper bound the ground truth such that it is not suitable to use as a lower bound. Compared to the high-precision singular value estimator, we see a noticeable difference to our low-precision estimator. It is worth noting that by definition our low-precision estimator upper bounds the high-precision one. Finally, we observe that all estimators follow similar trends, indicating that our estimator is suitable for optimization.
        \begin{figure}[tb]
            \footnotesize
            \centering
            	\includegraphics[width=\linewidth]{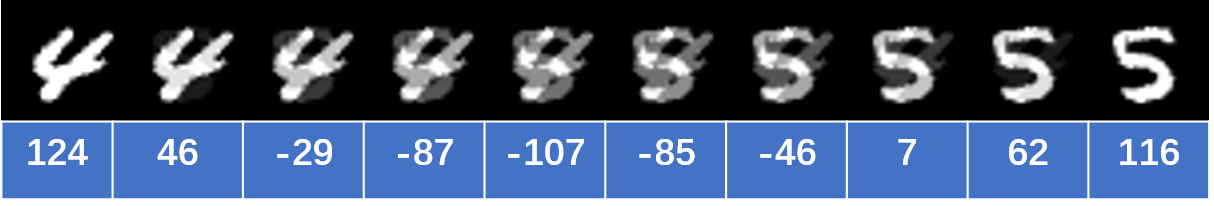} 
            	\\[-0.2em]
            	(a) $\EBM_{\SV+\GP}$
            	\\[0.5em]
            	\includegraphics[width=\linewidth]{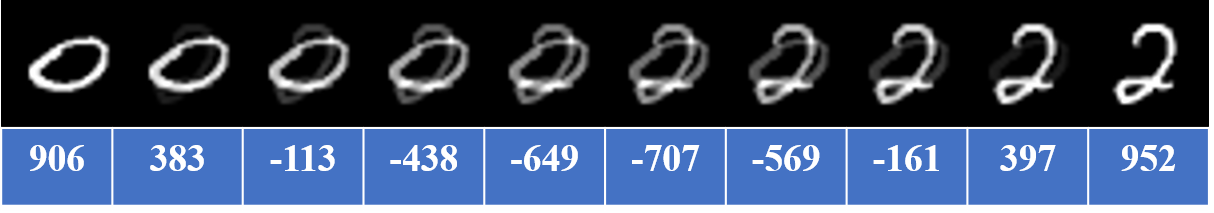} 
            	\\[-0.2em]
            	 (b) $\EBM_{\MI+\diff}$
            \caption{The negative energy of images generated by linear interpolation of two real images. Real images have higher negative energy than synthetic ones.}
            \label{mnist density} 
        \end{figure}

  \textbf{Test log-likelihood.~~}
        To compare different combinations of bounds, we consider an example where we can evaluate the exact log-likelihood associated with our energy function.
        To ensure that we can evaluate the exact log-likelihood, we use a normalizing flow \cite{dinh2014nice} for the energy function. In this case, for $\EBM_{\SV+\GP}$, $E_{\theta}(\vx)=-\log p_\theta(\vx)=-\log p_z(f_\theta(\vx))-\log \left(\left\vert\operatorname{det}\left(\frac{\partial f_\theta(\vx)}{\partial \vx}\right)\right\vert\right)$, where $f_\theta(\vx)$ is a invertible neural network. Likewise, for $\EBM_{\MI+\diff}$, $E_{\theta}(\vx,t)=-\log p_{\theta t}(\vx)=-\log p_z(f_\theta(\vx,t))-\log \left(\left\vert\operatorname{det}\left(\frac{\partial f_\theta(\vx,t)}{\partial \vx}\right)\right\vert\right)$. As discussed in \cref{sec:upper_bound2}, this need not be a valid bound as it may not satisfy the Fokker-Planck equation, but empirically we found it to be stable.
        We train six different models by considering all combinations of our proposed bounds as well as a zero-centered gradient penalty~($\ZGP$) as an approximate upper bound. The latter is included in the comparison as it is widely recognized as a common trick for stabilizing training, even though it does not constitute a bound. However, it exhibits instability in our experiment.
        \Cref{log-likelihood} shows test set log-likelihoods for the different bound combinations for training. We observe that except for the $\ZGP$-based upper bound, both
        the gradient penalty and score-based diffusion upper bound give satisfying results with both lower bounds. Among them, the singular value lower bound performs slightly better than the mutual information lower bound with both upper bounds. In fact, the heuristically defined zero-centered gradient penalty demonstrates instability during training when used with the singular value lower bound. Thus we set the number of inner loop iterations in $\SV$ to zero, which means a Hutchinson estimator for computing the entropy.
        
\textbf{Bidirectional bounds.~~}
        To verify the bidirectional characteristic of our method, we plot the upper and lower bound during training for each combination. As before, we use a normalizing flow model as an energy function. \Cref{bounds} show the results. Note that
        the diffusion-based upper bound depends on a constant that we cannot evaluate \eqref{improved upper bound}, which may explain why the upper bound in $\EBM_{\MI+\diff}$ and $\EBM_{\SV+\diff}$ are quite loose. We do not have a tightness measure for this upper bound.
        The gradient penalty upper bound equals the lower bound when $p_{\theta}(\vx)=p_g(\vx)$. Since we approximate the entropy used for computing the gradient penalties upper bound, we also cannot give an exact tightness measure here. We can, however, give the training trends for each combination of bidirectional bounds. In \cref{bounds}, we observe that, except for the baseline model without bidirectional bounds,
        all combinations of bidirectional bounds can give stable training curves, yielding decent generation results. No notable visualization differences in the generation performance between each combination.

        \textbf{Energy estimation.~~}
        To further verify if our energy can provide useful information about the density of data space, we linearly interpolate two randomly chosen images and compute the negative energy of the interpolated images. We use CNNs for both energy function $E_\theta$, generator $G$, and discriminator $T$. From \cref{mnist density} we observe that the real images have a higher likelihood under the energy functions than the synthetic (interpolated) images. This suggests that adversarial training pushes probability mass toward real images.\looseness=-1
        
    \subsection{Natural images}
        \noindent\textbf{Image generation.~~}
        So far we have only considered toy data, and do not challenge contemporary generative models, including our EBMs. Next we, therefore, consider more challenging benchmark data, specifically the $32 \times 32$ CIFAR-10~\citep{krizhevsky2009learning} dataset, the $64 \times 64$ cropped ANIMEFACE\footnote{https://www.kaggle.com/splcher/animefacedataset} dataset, and the $64 \times 64$ CelebA~\citep{liu2015deep} dataset. All represent a significant increase in complexity. For CIFAR-10 and ANIMEFACE datasets, we report Inception Score (IS), Fréchet Inception Distance (FID) score, and two $F_\beta$ scores \citep{sajjadi2018assessing} to measure model performance. \Cref{image generation} provides results for both our methods and the considered baselines, where we note that all adversarial-based EBMs and GANs use the same network architecture. We use the DCGAN~\citep{radford2015unsupervised} network architecture for CIFAR-10 and a Resnet architecture~\citep{kang2020ContraGAN} for ANIMEFACE. Hyper-parameters are chosen as in the original papers. We were unable to reproduce the reported results for VERA \citep{nomcmc} so we instead performed an extensive grid search to select the best hyper-parameters. Like the original paper, we choose the entropy weight to be $0.0001$. We relied on a public  implementation\footnote{https://github.com/rosinality/denoising-diffusion-pytorch} for DDPM. For EBM-Triangle, we reproduce results using the official code. For other models, we take the results from the original papers.
        \begin{figure*}[tb]
        	\footnotesize
        	\centering
        	\renewcommand{\tabcolsep}{1pt} \renewcommand{\arraystretch}{0.1} \begin{tabular}{ccccc}
        	\vspace{3pt}
        		\includegraphics[width=0.2\linewidth]{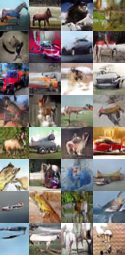} &
        		\includegraphics[width=0.2\linewidth]{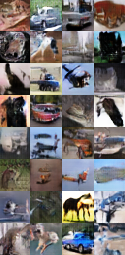} &
        	\includegraphics[width=0.2\linewidth]{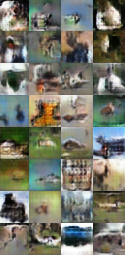} &
        		\includegraphics[width=0.2\linewidth]{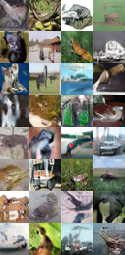}&
        	\includegraphics[width=0.2\linewidth]{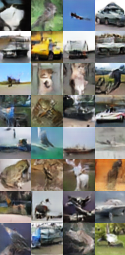}
        		\\ 
        		  (a) WGAN-0GP & (b) MEG & (c) VERA & (d) $\EBM_{\SV+\GP}$ & (e) $\EBM_{\MI+\diff}$ \\
        		 	\vspace{3pt}
        	\end{tabular}
        	\caption{
        	Generated samples on CIFAR-10 with our method and various methods.
        	}
        	\label{cifar10}
        \end{figure*}

        \begin{figure*}[tb]
        	\footnotesize
        	\centering
        	\renewcommand{\tabcolsep}{1pt} \renewcommand{\arraystretch}{0.1} \begin{tabular}{ccccc}
        	\vspace{3pt}
        		\includegraphics[width=0.2\linewidth]{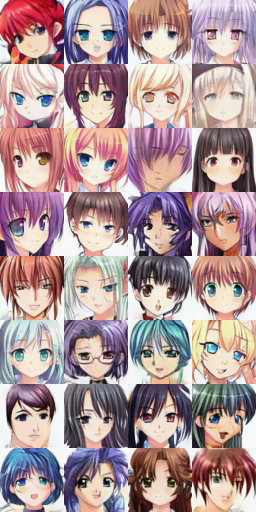} &
        		\includegraphics[width=0.2\linewidth]{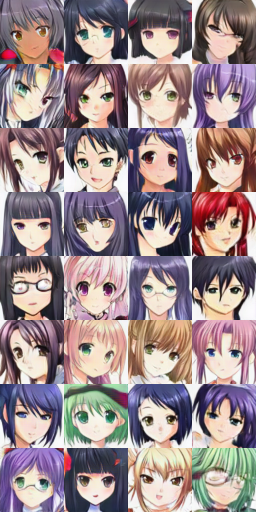} &
        		\includegraphics[width=0.2\linewidth]{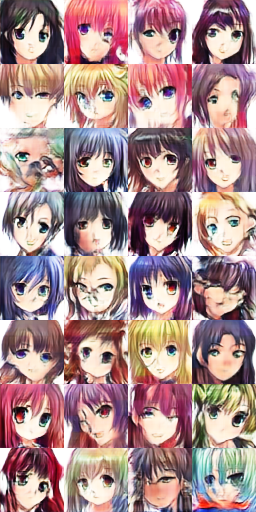}&
        		\includegraphics[width=0.2\linewidth]{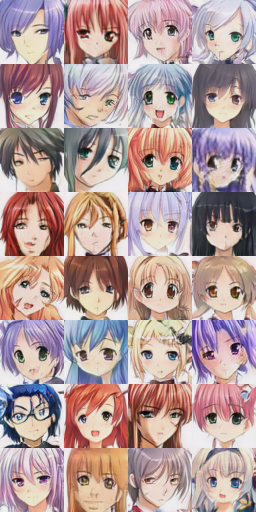} &
        		\includegraphics[width=0.2\linewidth]{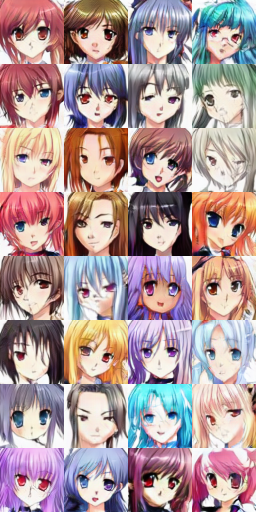}
        		\\
	       (a) WGAN-0GP & (b) MEG & (c) VERA & (d) $\EBM_{\SV+\GP}$ & (e) $\EBM_{\MI+\diff}$ \\
        		
        	\end{tabular}
        	\caption{
        	Generated samples on ANIMEFACE with our method and other generative models
        	}
        	\label{anemiface}
        \end{figure*}
        \Cref{image generation} shows that among the EBMs, our $\EBM_{\SV+\GP}$ generally performs the best on CIFAR-10, though $\EBM_{\MI+\diff}$ has the best IS score. 
        MEG, EGAN-Ent-VI, and VERA derive mutual information or variational approximations for the intractable entropy terms. EBM-Triangle and EBM-VAE use the joint distribution between latent and data space to avoid the intractable entropy term. But they all minimize a lower bound when optimizing the energy function.
        The EBMs are, however, generally outperformed by NCSN and DDPM, which is not surprising as NCSN and DDPM both use UNet architecture which is much bigger and more complex
        than our simple DCGAN architecture. To demonstrate this, we also scale up our network to UNet and get comparable results with NCSN and DDPM, see \cref{sec:cifar10 unet} for details. 
        On ANIMEFACE we see that our method is highly competitive. \Cref{cifar10,anemiface} show samples from different models. On ANIMEFACE, our EBMs yield more diverse samples with respect to face parts than the competing methods. All models mainly generate feminine faces. This could indicate that all models to some extent suffer from mode collapse, though our methods seem less sensitive to this issue. We further draw attention to the many corrupt samples drawn from VERA, despite extensive parameter search. In contrast, with CIFAR-10, we observed no notable differences in the generative capabilities between models.

    For CelebA dataset, we report the FID score in \Cref{celeba FID} since this measure metric is the most common score for comparison of current generative models. 
    We use 162,770 images for training and 19,867 for testing, the official split in PyTorch. The FID score is computed with testing images as the reference images.
    From \Cref{celeba FID} we can see our $\EBM_{\SV+\GP}$ get the best results among GANs and EBMs except for EBM-Diffusion. DDPM and EBM-Diffusion get the best FID score. Both methods use a diffusion process to learn data distribution, which can obtain the advantage of a diffusion-based model on generation quality. But they also have the disadvantage of being computationally expensive. Our $\EBM_{\MI+\diff}$ performs slightly worse on CelebA but still gets comparable results with other EBMs.

\textbf{Capacity usage.~~}
        In general, we want the proposal distribution $p_g$ to use as much of the capacity of the generator network $G$ as possible in order to be adaptive. To compare across models, we consider the anisotropy index \citep{wang2021geometry}, which is an implicit measure of capacity usage $\mathcal{C}_{\vz}$ that locally measures the intrinsic dimensionality of $G(\vz)$. Specifically, for a given $\vz$, this measures the standard deviation of the norm of the directional derivative of $G(\vz)$ with respect to the input dimensions, i.e.
         \begin{align}
         \begin{split}
            \mathcal{C}_{\vz} &= \mathrm{std}\left( \{ \| \mat{J}_{\vz} \vec{e}_i \| \}_{i = 1\ldots d} \right),
             \end{split}
        \end{align}
        where $\mathrm{std}(\cdot)$ denotes the standard deviation, and $\vec{e}_i$ is the $i^{\mathrm{th}}$ standard basis vector. Small values of $\mathcal{C}_{\vz}$ are indicative that all input dimensions contribute equally to the output of $G$, which suggests good capacity usage. We report the average value of $\mathcal{C}_{\vz}$ for $\vz \sim \mathcal{N}(\vec{0}, \mat{I}_d)$ across several runs in \cref{disentangling}. Most models, except VERA, perform well on CIFAR-10, while performance on ANIMEFACE is more diverse. In both cases, $\EBM_{\SV+\GP}$ and $\EBM_{\MI+\diff}$ show the most capacity usage. The significant difference between $\EBM_{\ZGP}$ and the bidirectional EBMs ($\EBM_{\SV+\GP}$ and $\EBM_{\MI+\diff}$) on ANIMEFACE suggests that the use of an upper bound helps to maximize the entropy of the proposal distribution $p_g$.
        \begin{table}[tb]
            \caption{Anisotropy indices ($\downarrow$).}
            \label{disentangling}
            \centering
            \begin{tabular}{clclcl}
            \toprule
            \multicolumn{2}{c}{Model} & \multicolumn{2}{c}{CIFAR-10} & \multicolumn{2}{c}{ANIMEFACE} \\ \midrule
            \multicolumn{2}{c}{WGAN-0GP} & \multicolumn{2}{c}{1.0823 $\pm$ 0.005} & \multicolumn{2}{c}{2.4366 $\pm$ 0.129} \\
            \multicolumn{2}{c}{MEG} & \multicolumn{2}{c}{0.9467 $\pm$ 0.009} & \multicolumn{2}{c}{2.3600 $\pm$ 0.142} \\
            \multicolumn{2}{c}{VERA} & \multicolumn{2}{c}{3.9694 $\pm$ 0.055} & \multicolumn{2}{c}{2.2929 $\pm$ 0.2074} \\
            \multicolumn{2}{c}{$\EBM_{\ZGP}$} & \multicolumn{2}{c}{1.0688 $\pm$ 0.0226} & \multicolumn{2}{c}{3.4560 $\pm$ 0.0521} \\
            \multicolumn{2}{c}{$\EBM_{\SV+\GP}$} & \multicolumn{2}{c}{\textbf{0.9431 $\pm$ 0.0191}} & \multicolumn{2}{c}{\textbf{1.8016 $\pm$ 0.1345}} \\
            \multicolumn{2}{c}{$\EBM_{\MI+\diff}$} & \multicolumn{2}{c}{1.0092 $\pm$ 0.0107} & \multicolumn{2}{c}{1.9067 $\pm$ 0.0617} 
            \\ \bottomrule
            \end{tabular}
        \end{table}
          \begin{table}[tb]
            \centering
            \caption{Comparison in terms of FID, Inception Score, $F_8$ score (weights recall higher than precision) and $F_{\sfrac{1}{8}}$ score (weights precision higher than recall).}
            \label{image generation}
            \resizebox{\linewidth}{!}{%
            \begin{tabular}{clclllclcl}
            \toprule
            \multicolumn{2}{c}{Model} & \multicolumn{2}{c}{Inception$\uparrow$} & \multicolumn{2}{c}{FID$\downarrow$} & \multicolumn{2}{c}{$F_8\!\!\uparrow$} & \multicolumn{2}{c}{$F_{\sfrac{1}{8}}\!\uparrow$} \\ \midrule
            \multicolumn{10}{c}{CIFAR-10} \\ \midrule
            \multicolumn{2}{c}{DCGAN} & \multicolumn{2}{c}{6.64} & \multicolumn{2}{c}{49.03} & \multicolumn{2}{c}{0.795} & \multicolumn{2}{c}{0.83} \\
            \multicolumn{2}{c}{WGAN-0GP} & \multicolumn{2}{c}{7.24 $\pm$ 0.035} & \multicolumn{2}{c}{29.31 $\pm$ 0.185} & \multicolumn{2}{c}{0.92 $\pm$ 0.010} & \multicolumn{2}{c}{0.95 $\pm$ 0.010} \\
            \multicolumn{2}{c}{CoopNets~\citeyearpar{xie2018cooperative}} & \multicolumn{2}{c}{6.55} & \multicolumn{2}{c}{36.4} & \multicolumn{2}{c}{-} & \multicolumn{2}{c}{-} \\
            \multicolumn{2}{c}{CoopNets~\citeyearpar{xie2018cooperative2}} & \multicolumn{2}{c}{-} & \multicolumn{2}{c}{33.61} & \multicolumn{2}{c}{-} & \multicolumn{2}{c}{-} \\
            \multicolumn{2}{c}
            {EGAN-Ent-VI~\citep{dai2017calibrating}} & \multicolumn{2}{c}{7.07 $\pm$ 0.10} & \multicolumn{2}{c}{-} & \multicolumn{2}{c}{-} & \multicolumn{2}{c}{-} \\
            \multicolumn{2}{c}{EBM-VAE~\citep{xie2021learning}} & \multicolumn{2}{c}{6.65} & \multicolumn{2}{c}{36.2} & \multicolumn{2}{c}{-} & \multicolumn{2}{c}{-} \\
            \multicolumn{2}{c}{EBM-Triangle~\citep{han2019divergence}} & \multicolumn{2}{c}{7.30} & \multicolumn{2}{c}{28.96} & \multicolumn{2}{c}{0.89} & \multicolumn{2}{c}{0.96} \\
            \multicolumn{2}{c}{IGEBM~\citep{du2019implicit}} & \multicolumn{2}{c}{6.78} & \multicolumn{2}{c}{38.2} & \multicolumn{2}{c}{-} & \multicolumn{2}{c}{-} \\
            \multicolumn{2}{c}{Short-Run MCMC~\citep{nijkamp2019learning}} & \multicolumn{2}{c}{6.21} & \multicolumn{2}{c}{44.50} & \multicolumn{2}{c}{-} & \multicolumn{2}{c}{-} \\
            \multicolumn{2}{c}{MEG} & \multicolumn{2}{c}{6.62 $\pm$ 0.243} & \multicolumn{2}{c}{34.55 $\pm$ 1.145} & \multicolumn{2}{c}{0.88 $\pm$ 0.001} & \multicolumn{2}{c}{0.92 $\pm$ 0.010} \\
            \multicolumn{2}{c}{VERA} & \multicolumn{2}{c}{5.06 $\pm$  0.555} & \multicolumn{2}{c}{66.38 $\pm$ 6.635} & \multicolumn{2}{c}{0.58 $\pm$ 0.080} & \multicolumn{2}{c}{0.79 $\pm$ 0.005} \\
            \multicolumn{2}{c}{NCSN} & \multicolumn{2}{c}{8.87 $\pm$  0.12} & \multicolumn{2}{c}{25.32} & \multicolumn{2}{c}{-} & \multicolumn{2}{c}{-} \\
            \multicolumn{2}{c}{DDPM} & \multicolumn{2}{c}{\textbf{9.03}} & \multicolumn{2}{c}{\textbf{7.76}} & \multicolumn{2}{c}{\textbf{0.98}} & \multicolumn{2}{c}{\textbf{0.99}} \\
            \multicolumn{2}{c}{$\EBM_{\ZGP}$} & \multicolumn{2}{c}{6.90 $\pm$ 0.032} & \multicolumn{2}{c}{35.42 $\pm$ 0.582} & \multicolumn{2}{c}{0.90 $\pm$ 0.004} & \multicolumn{2}{c}{0.93 $\pm$ 0.002} \\
            \multicolumn{2}{c}{$\EBM_{\SV+\GP}$} & \multicolumn{2}{c}{7.45 $\pm$ 0.014} & \multicolumn{2}{c}{28.63 $\pm$ 0.290} & \multicolumn{2}{c}{0.93 $\pm$ 0.001} & \multicolumn{2}{c}{0.95 $\pm$ 0.008} \\ 
            \multicolumn{2}{c}{$\EBM_{\MI+\diff}$} & \multicolumn{2}{c}{7.46$\pm$0.051} & \multicolumn{2}{c}{30.37 $\pm$ 0.054} & \multicolumn{2}{c}{0.92 $\pm$ 0.003} & \multicolumn{2}{c}{0.95 $\pm$ 0.004} \\ 
            \midrule
            \multicolumn{10}{c}{ANIMEFACE} \\ \midrule
            \multicolumn{2}{c}{WGAN-0GP} & \multicolumn{2}{c}{2.22 $\pm$ 0.030} & \multicolumn{2}{c}{9.76 $\pm$ 0.674} & \multicolumn{2}{c}{\textbf{0.95 $\pm$ 0.005}} & \multicolumn{2}{c}{\textbf{0.98 $\pm$ 0.005}} \\
            \multicolumn{2}{c}{MEG} & \multicolumn{2}{c}{2.20 $\pm$ 0.020} & \multicolumn{2}{c}{9.31 $\pm$ 0.007} & \multicolumn{2}{c}{\textbf{0.95 $\pm$ 0.005}} & \multicolumn{2}{c}{\textbf{0.98 $\pm$ 0.001}} \\
            \multicolumn{2}{c}{VERA} & \multicolumn{2}{c}{2.15 $\pm $ 0.001} & \multicolumn{2}{c}{41.00 $\pm$ 1.072} & \multicolumn{2}{c}{0.515 $\pm$ 0.078} & \multicolumn{2}{c}{0.78 $\pm$ 0.013} \\
            \multicolumn{2}{c}{DDPM} & \multicolumn{2}{c}{2.18} & \multicolumn{2}{c}{\textbf{8.81}} & \multicolumn{2}{c}{0.94} & \multicolumn{2}{c}{\textbf{0.98}} \\
            \multicolumn{2}{c}{$\EBM_{\ZGP}$} & \multicolumn{2}{c}{\textbf{2.26 $\pm$ 0.017}} & \multicolumn{2}{c}{20.53 $\pm$ 0.524} & \multicolumn{2}{c}{0.889 $\pm$ 0.008} & \multicolumn{2}{c}{0.909 $\pm$ 0.019} \\
            \multicolumn{2}{c}{$\EBM_{\SV+\GP}$} & \multicolumn{2}{c}{\textbf{2.26 $\pm$ 0.005}} & \multicolumn{2}{c}{12.75 $\pm$ 0.045} & \multicolumn{2}{c}{0.94 $\pm$ 0.001} & \multicolumn{2}{c}{0.96 $\pm$ 0.005} 
            \\ 
             \multicolumn{2}{c}{$\EBM_{\MI+\diff}$} & \multicolumn{2}{c}{2.17$\pm$0.065} & \multicolumn{2}{c}{13.32$\pm$0.45} & \multicolumn{2}{c}{0.92$\pm$0.001} & \multicolumn{2}{c}{0.97$\pm$0.004} \\
            \bottomrule
            \end{tabular}}
        \end{table}
\begin{table}[htbp]
\centering
\caption{FID scores on $64\times 64$ CelebA dataset}
\label{celeba FID}
\resizebox{0.4\textwidth}{!}{%
\begin{tabular}{cc}
\hline
\tiny
Model & FID \\ \hline
WGAN-0GP & \multicolumn{1}{l}{8.59 $\pm$ 0.179} \\
MEG & \multicolumn{1}{l}{8.52 $\pm$ 0.042} \\
Joint-EBM-Triangle~\citep{han2020joint} & 24.7 \\
BiDVL~\citep{kan2022bi} & 17.24 \\
EBM-Diffusion~\citep{gao2020learning} & 5.98 \\
DDPM & \textbf{3.50} \\
$\EBM_{\ZGP}$ & \multicolumn{1}{l}{10.72 $\pm$ 0.314} \\
$\EBM_{\SV+\GP}$ & \multicolumn{1}{l}{8.39 $\pm$ 0.567} \\
$\EBM_{\MI+\diff}$ & 14.79$\pm$1.08 \\ \hline
\end{tabular}%
}
\end{table}

        \textbf{Out-of-distribution detection.~~}
        In principle, a well-estimated EBM should assign a lower likelihood to out-of-distribution (OOD) observations than to observations from the training distribution. OOD detection performance based on likelihood can then serve as a measure of density quality. We compare the relevant baselines following established practice \citep{hendrycks2016baseline, hendrycks2018deep, alemi2018uncertainty, choi2018waic, ren2019likelihood, havtorn2021hierarchical} and report the threshold independent evaluation metrics of Area Under the Receiver Operator Characteristic (AUROC$\uparrow$), Area Under the Precision-Recall Curve (AUPRC$\uparrow$) and False Positive Rate at 80\% (FPR80$\downarrow$). Here the arrow indicates the direction of improvement of the metrics. \Cref{OOD} provide the results, where each column takes the form `In-distribution / Out-of-distribution' in reference to the training and test set, respectively.
        On CIFAR-10~/~SVHN we note that $\EBM_{\SV+\GP}$ works the best. On  CIFAR-10~/~CIFAR-100 we generally observed poorer results, but $\EBM_{\ZGP}$ did quite a bit better than the rest. The general degradation is most likely due to the strong similarity between CIFAR-10 and CIFAR-100. For ANIMEFACE~/~Bedroom, however, a similar situation occurs, even if the datasets are not alike. Here $\EBM_{\MI+\diff}$ is significantly better than the rest.\looseness=-1
        
        This study provides no clear winner. Our approach consistently performs well, but perhaps surprisingly, this is also the case for WGAN-0GP, which is not a density estimator.OOD detection is a task known for its subtle pitfalls~\citep{havtorn2021hierarchical} and high variance across training stages in EBMs~\citep{kan2022bi},
        so we argue that the presented results should be taken with a grain of salt even if our model performs well.
        
        \begin{table*}[tb]
            \centering
            \caption{AUROC$\uparrow$, AUPRC$\uparrow$, and FPR80$\downarrow$ for OOD detection for `train/test' datasets.}
            \label{OOD}
            \resizebox{\textwidth}{!}{%
            \centering
            \begin{tabular}{cccccccccc}
            \toprule
            \multirow{2}{*}{Model} & \multicolumn{3}{c}{CIFAR-10~/~SVHN} & \multicolumn{3}{c}{CIFAR-10~/~CIFAR-100} & \multicolumn{3}{c}{ANIMEFACE~/~Bedroom} \\ \cmidrule(lr){2-4}\cmidrule(lr){5-7}\cmidrule(lr){8-10}
             & AUROC$\uparrow$ & AUPRC$\uparrow$ & FPR80$\downarrow$ & AUROC$\uparrow$ & AUPRC$\uparrow$ & FPR80$\downarrow$ & AUROC$\uparrow$ & AUPRC$\uparrow$ & FPR80$\downarrow$ \\ \midrule
            WGAN-0GP & 0.8 & 0.83 & 0.39 & 0.54 & 0.55 & 0.77 & 0.51 & 0.48 & 0.72 \\
            MEG & 0.79 & 0.81 & 0.42 & 0.52 & 0.53 & 0.8 & 0.56 & 0.53 & 0.77 \\
            VERA & 0.62 & 0.64 & 0.64 & 0.51 & 0.51 & 0.79 & 0.60 & 0.525 & 0.6 \\
            $\EBM_{\ZGP}$~(ours) & 0.66 & 0.69 & 0.67 & \textbf{0.64} & \textbf{0.63} & \textbf{0.64} & 0.67 & 0.63 & 0.53 \\
            $\EBM_{\SV+\GP}$~(ours) & \textbf{0.88} & \textbf{0.86} & \textbf{0.1997} & 0.53 & 0.52 & 0.7765 & 0.53 & 0.505 & 0.83 \\ 
            $\EBM_{\MI+\diff}$~(ours) & 0.53 & 0.52 & 0.73 & 0.52 & 0.51 & 0.78 & \textbf{0.84} & \textbf{0.72} & \textbf{0.201} \\ 
            \bottomrule
            \end{tabular}
            }
        \end{table*}
        \textbf{Comparison of different combinations of bounds.~~}
        To explore the performance of different combinations of upper and lower bounds on natural images, we do a comparison with different bounds on CIFAR-10 across different metrics. \Cref{comparison bounds} shows the results. We do not see a clear winner in this comparison.
        For the computational cost, $EBM_{\MI+\diff}$ is time efficient and is around 1.3$\times$ of WGAN-0GP, other types of our bidirectional EBMs require to compute the smallest singular value and are around 2.5$\times$ of baseline. $\EBM_{\SV+\GP}$ and $\EBM_{\SV+\diff}$ have no extra network and are space-efficient. Hence, the selection of bidirectional bounds depends on training requirements, such as finding the optimal balance between memory usage and training time.
        
        \textbf{Robustness of upper bound.~~}
        We next test the robustness of our two upper bounds to changes in parameters. As described in \cref{training details}, when using the singular value-based lower bound on large-scale datasets we reduce the weight of entropy regularizer $\lambda$ to improve performance. We combine this lower bound with our two upper bounds respectively and consider different values of $\lambda$. Results on CIFAR-10 are in \cref{robustness}. We see that with the diffusion-based bound we always get satisfying performance, while the gradient penalty bound results in a loss in performance when $\lambda = 1$.
        \begin{table}[tb]
            \caption{Comparison with combinations of different upper and lower bounds on the CIFAR-10 dataset.}
            \label{comparison bounds}
            \centering
            \resizebox{\linewidth}{!}{%
            \begin{tabular}{ccccc}
            \toprule
            \multirow{2}{*}{Metric} & \multicolumn{2}{c}{$\GP$} & \multicolumn{2}{c}{$\diff$} \\ \cmidrule(lr){2-3} \cmidrule(lr){4-5}
             & \multicolumn{1}{c}{$\MI$} & $\SV$ & \multicolumn{1}{c}{$\MI$} & $\SV$ \\ \midrule
            FID$\downarrow$ & \multicolumn{1}{c}{28.96$\pm$1.251} & \textbf{28.63 $\pm$ 0.290} & \multicolumn{1}{c}{30.37$\pm$0.054} & 29.31$\pm$0.261 \\
            IS$\uparrow$ & \multicolumn{1}{c}{7.41$\pm$0.093} & 7.45 $\pm$ 0.014 & \multicolumn{1}{c}{7.46$\pm$0.051} & \textbf{7.65$\pm$0.008} \\
            F8$\uparrow$ & \multicolumn{1}{c}{0.92$\pm$0.004} & \textbf{0.93 $\pm$ 0.001} & \multicolumn{1}{c}{0.92$\pm$0.003} & 0.93$\pm$0.004 \\
            F1/8$\uparrow$ & \multicolumn{1}{c}{\textbf{0.96$\pm$0.003}} & 0.95 $\pm$ 0.008 & \multicolumn{1}{c}{0.95$\pm$0.004} & \multicolumn{1}{c}{0.95$\pm$0.005} \\ \bottomrule
            \end{tabular}}
        \end{table}
        \begin{table*}[tb]
            \centering
            \caption{Robustness of upper bounds with different entropy regularizer weight $\lambda$.}
            \label{robustness}
            \begin{tabular}{ccccccc}
            \toprule
             & \multicolumn{3}{c}{$\diff$} & \multicolumn{3}{c}{$\GP$} \\ \cmidrule(lr){2-4} \cmidrule(lr){5-7}
            $\lambda$ & 0.0001 & 0.01 & 1 & 0.0001 & 0.01 & 1 \\ \midrule
            FID$\downarrow$ & 29.31$\pm$0.261 & 30.4$\pm$0.65 & 33.13$\pm$0.349 & 28.63$\pm$0.290 & 30.38$\pm$0.22 & 45.08$\pm$0.285 \\
            IS$\uparrow$ & 7.65$\pm$0.008 & 7.35$\pm$0.079 & 7.12$\pm$0.079 & 7.45$\pm$0.014 & 7.17$\pm$0.085 & 6.16$\pm$0.126 \\
            F8$\uparrow$ & 0.931$\pm$0.004 & 0.906$\pm$0.006 & 0.903$\pm$0.002 & 0.93$\pm$0.001 & 0.91$\pm$0.002 & 0.78$\pm$0.007 \\
            F1/8$\uparrow$ & 0.95$\pm$0.005 & 0.956$\pm$0.000 & 0.94$\pm$0.004 & 0.95$\pm$0.008 & 0.95$\pm$0.002 & 0.91$\pm$0.006 \\ \bottomrule
            \end{tabular}%
        \end{table*}
  \section{Discussion}
    This paper stated from the observation that current GAN-like methods for training energy-based models (EBMs) both minimize and maximize a lower bound. We have put forward the hypothesis that this may cause training instability, and have instead proposed to bound the negative log-likelihood from above and below and switch between bounds when minimizing and maximizing. We then propose a series of such bounds and compare their empirical behavior. The lower bound based on singular values is similar to existing ones, but the proposed algorithm for its evaluation is new. Unlike previous methods, this new algorithm does not require any additional networks and is generally easy to tune; the only parameter is the number of iterations, which reflects a trade-off between computational cost and the accuracy of the bound.
    The mutual information-based lower bound needs an extra network but it is, in turn, time efficient. The two lower bounds then respectively give space or time efficiency.
    
    The idea of introducing an upper bound, $\ceil{\L}$, is new to the literature. Our gradient penalty-based bound is, however, similar to commonly called upon heuristics to stabilize training. To the best of our knowledge, this is the first time that gradient penalties have been shown to provide a bound of the log-likelihood. We consider it rather neat that current engineering practice, thus, can be justified directly from a probabilistic perspective.

    The general empirical finding across datasets and tasks is that our bidirectional bounds generally lead to models that perform as well or better than some of the current state-of-the-art. The evidence suggests that the use of bidirectional bounds causes the generator to increase both its entropy and its capacity usage. We observe this both through direct measurement and improved sample quality.\looseness=-1 
    
 \textbf{Limitations.~~}
    We mainly have four concerns with the proposed methods. 
    First, for our score-based diffusion upper bound, it is not clear how to design $E_{\theta}(\x,t)$ to make it satisfy the Fokker-Planck equation, and this part remains heuristic.
    Second, our lower bound based on singular values relies on an iterative solver. While our implementation is quite efficient, it must still run for several iterations to ensure convergence. Our practical implementation, therefore, stops prematurely to ensure fast computations. We found this to empirically work well even if it technically violates the bound. 
    Furthermore, to get good performance we need a weight regularizer $\lambda$ to balance the entropy term in the lower bound based on singular values. In general, this weight regularizer violates the lower bound, but in practice, we almost always have positive entropies, in which case the bound remains valid.
    Finally, for the gradient penalty upper bound, we need access to the volume of support $M$ for the proposal distribution $p_g$. We do not have a method to compute this and therefore consider the ratio $\frac{M}{s_1^2}$ between $M$ and the square of the estimated smallest singular value $s_1^2$  as a parameter that is either fixed or dynamic decay. When used in conjunction with the singular value lower bound, we observe that $\frac{M}{s_1^2}$ needs to be adjusted with the weight regularizer $\lambda$ to get a satisfying performance.
\section*{Declarations}
\subsection{Funding}
This work was supported by the National Natural Science Foundation of China under Grant U24A20220 and partly funded by the Novo Nordisk Foundation through the Center for Basic Machine Learning Research in Life Science (NNF20OC0062606), STCSM under Grant 22DZ2229005, and 111 project BP0719010. It also received funding from the European Research Council (ERC) under the European Union’s Horizon 2020 research, innovation program (757360), National science foundation of China under grant 61771305, and Shanghai Municipal Science and Technology Major Project (2021SHZDZX0102). JF was supported in part by the Novo Nordisk Foundation (NNF20OC0065611) and the Independent Research Fund Denmark (9131-00082B). SH was supported in part by research grants (15334, 42062) from VILLUM FONDEN. The authors also acknowledge the support of the Pioneer Centre for AI, DNRF grant number P1.
\subsection{Competing interests}
\begin{itemize}
    \item The authors have no relevant financial or non-financial interests to disclose.
    \item The authors have no competing interests to declare that are relevant to the content of this article.
    \item All authors certify that they have no affiliations with or involvement in any organization or entity with any financial interest or non-financial interest in the subject matter or materials discussed in this manuscript.
    \item The authors have no financial or proprietary interests in any material discussed in this article.
\end{itemize}
\subsection{Data availability}
All data analyzed during this study are publicly accessible.
\begin{appendices}

\section{Model architecture}
\label{model architecture}
Our key motivation for design of $p_g$ is to ensure that the model has support over the entire data space. This is needed for the reason that the KL divergence requires the two distributions to be defined on the same measurable space. Since our energy distribution is defined for the entire data space, we need the generated distribution to have the same support.  The simplest way to achieve this is with our choice of $x=G(z)+\epsilon$, where $\epsilon$ is a small-variance Gaussian variable. This concentrates the distribution around the manifold spanned by our generator while ensuring that the generated distribution has full support over the data space.
Moreover, if we do not add noise to our generator, our mutual information lower bound is infinite. The added noise makes the bound finite.

We consistently assume that $G: \mathbb{R}^d \rightarrow \mathbb{R}^D$ spans an immersed $d$-dimensional manifold in $\mathbb{R}^D$ as this allows us to apply the change-of-variables formula to get a density for the generator. In order to ensure that this assumption is valid, we minimally restrict the architecture of the generator neural network.

An immersion implies that the Jacobian of $G$ exists and has full rank. We can then ensure existence by using activation functions with at least one derivative almost everywhere. Any smooth activation satisfies this assumption, but we emphasize that, for example, ReLU activations lack a derivative at a single point. The non-smooth region has measure zero if the linear map inside the activation is not degenerate, in which case the change-of-variables technique still applies.

We are unaware of architectural tricks to ensure that the Jacobian has full rank. Still, a minimum requirement is no hidden layer may have dimensionality less than the $d$ dimensions of the latent space. This is a natural requirement that is practically always satisfied. Our model maximizes the generator entropy, which practically ensures full-rank Jacobians (noting that degeneracy in the Jacobian reduces the entropy). We have in practice generally not experienced degenerate Jacobians during training.
\begin{table*}
\centering
\caption{Selection of most important hyper-parameters and their
setting.}
\label{parameter setting}
\resizebox{\linewidth}{!}{%
\begin{tabular}{cccccc}
\toprule
Datasets & Optimization & \multicolumn{1}{l}{Learning rate} & \multicolumn{1}{l}{Batch size} & \multicolumn{1}{l}{Iterations/Epochs} & \multicolumn{1}{l}{latent dim} \\ \midrule
Toy & Adam(0.0,0.9) & 2e-4 & 200 & 150000 & 2 \\
MNIST & Adam(0.0,0.9) & 2e-4 & 100 & 90(epochs) & 32 \\
CIFAR-10 & Adam(0.0,0.999) & 5e-5 & 64 & 200000 & 128 \\
ANIMEFACE & Adam(0.0,0.999) & 5e-5 & 64 & 100000 & 128 \\ 
CelebA & Adam(0.0,0.999) & 5e-5 & 64 & 200000 & 128 \\ 
\bottomrule
\end{tabular}}
\end{table*}
\section{Training details}
\Cref{parameter setting} gives the hyper-parameters used during training. In \cref{training upper bound} we set $p=2$. We normalize the data to $[-1, 1]$ and do not use dequantization. We augment only using random horizontal flips during training. We used the official train-test split from PyTorch for cifar10 and an 85:15 train-test split for animeface. We set $\frac{M}{s_1^2}=\frac{0.1}{z_{dim}}$ for our gradient penalty upper bound, where $s_1$ is the smallest singular value of Jacobian and $z_{dim}$ is the latent dimension. We find this set of parameters generally works well across datasets. Furthermore, we choose $\frac{M}{s_1^2}$ with dynamic decay, i.e.\@ $\frac{M}{s_1^2}=\frac{a(b+i)^{-\gamma}}{2}$ where $i$ denotes the iteration, $\gamma = .55$ and $a$ and $b$ are chosen such that $\frac{2M}{s_1^2}$ decreases from $0.01$ to $0.0001$ during training. These dynamics are consistent with the approximation of $\KL{p_g}{p_\theta}$ from the Taylor series point of view. We find that practically for $\EBM_{\SV+\GP}$, $\frac{M}{s_1^2}$ must be adjusted with entropy regularizer $\lambda$ together. 
We also observe the implications of the network design on the final performance. For example, not using batch normalization in the energy function can improve the stability of training on ANIMEFACE dataset. 
\section{Selection of bidirectional bound pairings}
Given the two available options for both the lower and the upper bound, we provide recommendations for selecting bidirectional bound pairings based on our training experience. 
For real-world applications, as elaborated in the main text, we generally
recommend using combinations $\EBM_{\SV+\GP}$ or $\EBM_{\MI+\diff}$ first due to their respective advantages in memory and time efficiency.
If the data and latent space dimensions are relatively small and the network is shallow, we recommend using $\EBM_{\SV+\GP}$, as it tends to achieve better performance in such scenarios. Conversely, for larger dimensions or deeper networks, we suggest using $\EBM_{\MI+\diff}$ to reduce computational cost. Furthermore, $\GP$ and $\SV$ may suffer from training instability due to their reliance on iterative optimization. Therefore, in cases of unstable training, it is advisable to consider replacing $\GP$ or $\SV$ with $\diff$ or $\MI$.
      
\section{Additional results}
\subsection{CIFAR-10}
\label{sec:cifar10 unet}
Although our bidirectional EBM obtains superior performance among EBMs, it still lags far behind diffusion-based models in terms of generation metrics.
To demonstrate the scalability and potential of our model, we also scale up our network structure to UNet for CIFAR-10 dataset. We adopt the same architecture in UOTM~\citep{choi2024generative}, where its discriminator is utilized as our energy function. Our generator has the same scale as the network in NCSN and DDPM. We borrow some training strategies from UOTM to stabilize training and improve generative performance, such as the form of energy discrepancy and cost function. 
For the choice of bidirectional bounds, we only use our $\EBM_{\MI+\diff}$ since UNet network is complex and the input has the same dimension as the output, causing a costly Jacobian computation. We present our quantitative results in Table~\ref{CIFAR-10 UNet} and qualitative results in Fig.~\ref{CIFAR-10 UNet}.
 All baseline results used for comparison are sourced  from~\citet{xiao2021tackling} and \citet{song2023consistency}. For our method, we run the experiments with different seeds and report the standard deviation. 
For FID evaluation, we use the training set as a reference, aligning common practice in diffusion-based models~\citep{xiao2021tackling}.
From Table~\ref{CIFAR-10 UNet} we can observe our $\EBM_{\MI+\diff}$ achieves results comparable to diffusion-based models while using only one-step generation, offering significantly greater efficiency. Although recent techniques like distillation and consistency models can also enable one-step generation with minimal performance decline, our model demonstrates competitive results compared with them. This experiment suggests that the superiority of diffusion-based models stems more from the large network architecture and meticulous parameter fine-tuning than from the methodology itself. Our bidirectional EBM can achieve performance comparable to recent advanced diffusion-based models when equipped with a similar-scale UNet and an appropriate training strategy. Moreover,  as shown in~\cite{geng2024improving}, EBM can be further improved by integrating it with a diffusion process, which presents an opportunity for future work.
\begin{table}[htbp]
	\centering
	\caption{Comparison with diffusion-based models on $32\times 32$ CIFAR-10 dataset.}
	\label{CIFAR-10 UNet}
	\renewcommand{\arraystretch}{1}
    \resizebox{0.45\textwidth}{!}{%
		\begin{tabular}{cccc}
			\toprule
				Model          & FID$\downarrow$   & IS$\uparrow$   & NFE$\downarrow$  \\ \midrule
			
			DDPM~\cite{ho2020denoising}        & 3.21  & 9.46 & 1000   \\
			NCSN~\cite{song2019generative}          & 25.3  & 8.87 & 1000   \\
			DDIM~\cite{song2020denoising}     & 4.67 & 8.78 & 50  \\
			Score SDE~(VE)~\cite{song2020score}       & \textbf{2.20}  & 9.89 & 2000  \\
			Score SDE~(VP)~\cite{song2020score}     & 2.41 & 9.68  & 2000   \\
			DDPM Distillation~\cite{luhman2021knowledge}        & 9.36  & 8.36 & 1   \\
            CD~\citep{song2023consistency}              & 3.55  & 9.48 & 1 \\
            CT~\citep{song2023consistency}              & 8.70  & 8.49 & 1 \\
				$\EBM_{\MI+\diff}$~(Large) & 5.47 $\pm$ 0.14  & \textbf{10.20$\pm$0.09} & 1 \\ 
			\bottomrule
		\end{tabular}}
\end{table} 
\begin{figure}[htbp]
            \footnotesize
            \centering
            	\includegraphics[width=0.9\linewidth]{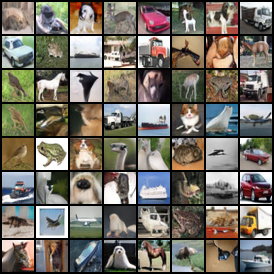} 
            	
            \caption{Visual results of CIFAR-10 using UNET.}
            \label{generation_church} 
        \end{figure}
\begin{figure*}[tb]
            \footnotesize
            \centering
            	\includegraphics[width=\linewidth]{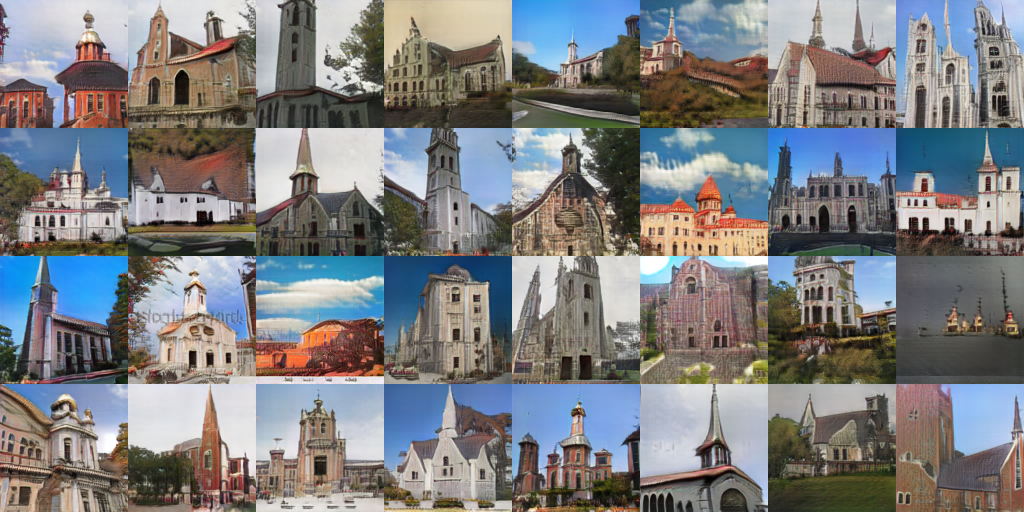} 
            	\\[-0.2em]
            	(a) $\EBM_{\SV+\GP}$
            	\\[0.5em]
            	\includegraphics[width=\linewidth]{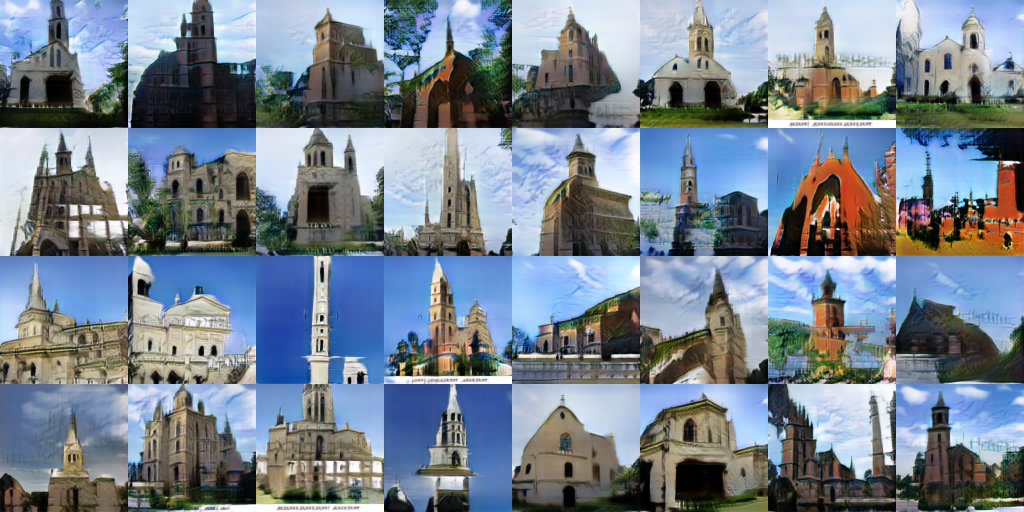} 
            	\\[-0.2em]
            	 (b) $\EBM_{\MI+\diff}$
            \caption{Visual results of $128\times128$ LSUN Church with two different settings.}
            \label{generation_church} 
        \end{figure*}
\subsection{Imagenet}
\begin{figure*}[t]
            \footnotesize
            \centering
            \renewcommand{\tabcolsep}{1pt} \renewcommand{\arraystretch}{0.1} \begin{tabular}{ccc}
            \includegraphics[width=0.33\linewidth]{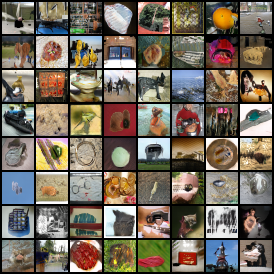}&
            \includegraphics[width=0.33\linewidth]{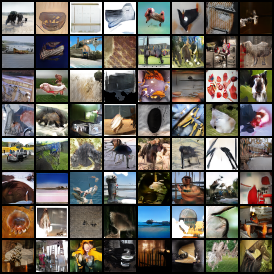}&
            \includegraphics[width=0.33\linewidth]{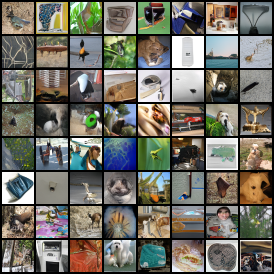} 
            	\vspace{1mm}\\
            (a) $\EBM_{\SV+\GP}$ &	(b) $\EBM_{\MI+\GP}$ & (c) $\EBM_{\MI+\diff}$~(Large)
             \end{tabular}
            \caption{Visual results of $32\times 32$ ImageNet dataset with different settings.
              }
            \label{visual imagenet} 
        \end{figure*}

To further evaluate the efficacy of our model on complex datasets that better approximate real-world scenarios, we assess its generative performance on $32\times 32$ ImageNet dataset, which contains 1,281,167 training images and 50,000 test images.
We perform experiments using both DCGAN and UNet architectures, as applied to the CIFAR-10 dataset. For  $\EBM_{\SV+\GP}$, we set the number of inner loop iterations in $\SV$ to zero and find that they contribute to stabilizing the training process; a detailed exploration of this phenomenon is left for future work. Table~\ref{Imagenet FID} and Fig.~\ref{visual imagenet} show our quantitative and qualitative results respectively.
For quantitative evaluation, we report only the FID score, as it is the most widely used metric for assessing generative models on this dataset.
From Table~\ref{Imagenet FID} we can see our $\EBM_{\MI+\diff}$-Large gets the best performance compared to other advanced generative models including diffusion-based model DDPM++~\citep{kim2021soft} and the latest energy-based model CDRL~\citep{zhu2023learning}. Even our $\EBM_{\SV+\GP}$ and $\EBM_{\MI+\diff}$ only use simple DCGAN architecture, they still outperform competitive EBM framework CLEL-Large which employs a deep ResNet network.
This experiment further verifies the effectiveness of our bidirectional EBM on large complex datasets. 
\begin{table}[tb]
\centering
\caption{FID scores on $32\times 32$ ImageNet dataset}
\label{Imagenet FID}
\resizebox{0.4\textwidth}{!}{%
\begin{tabular}{cc}
\hline
Model & FID \\ \hline
EBM-IG~\citep{du2019implicit} & \multicolumn{1}{c}{62.23} \\
PixelCNN~\citep{Salimans_Karpathy_Chen_Kingma_2017} & \multicolumn{1}{c}{40.51} \\
CLEL-Large~\citep{lee2023guiding} & \multicolumn{1}{c}{15.47} \\
DDPM++(VP, NLL)~\citep{kim2021soft} & \multicolumn{1}{c}{8.42} \\
CDRL~\citep{zhu2023learning} & \multicolumn{1}{c}{9.35} \\
$\EBM_{\SV+\GP}$ & \multicolumn{1}{l}{11.46 $\pm$ 0.236} \\
$\EBM_{\MI+\diff}$ & \multicolumn{1}{l}{14.61 $\pm$ 0.371} \\
$\EBM_{\MI+\diff}$~(Large) & \multicolumn{1}{l}{\textbf{6.57 $\pm$ 0.114}} \\
\hline
\end{tabular}%
}
\end{table}
 \subsection{LSUN Church}
We also test our model on a large-scale LSUN dataset. For this dataset, we choose the Church outdoor category with a size of $128\times 128$. We use 126,227 training images and 300 testing images following the official PyTorch split. The latent dimension is set to 256. We employ the same ResNet architecture as that used for the ANIMEFACE and CelebA datasets. We trained  $\EBM_{\SV+\GP}$ for 100,000 iterations and $\EBM_{\MI+\diff}$ for 200,000 iterations.
\cref{generation_church} illustrates the scalability of our models. While these samples are not state-of-the-art, they do showcase the ability of our model to scale to high-dimensional problems. Further improving these results most likely requires work on picking the right network architecture, fine-tuning parameters, and specialized optimization strategies. We leave this for future work.
\end{appendices}
\newpage
\bibliographystyle{plainnat}
\bibliography{bibliography}

\begin{thebibliography}{89}
\providecommand{\natexlab}[1]{#1}
\providecommand{\url}[1]{\texttt{#1}}
\expandafter\ifx\csname urlstyle\endcsname\relax
  \providecommand{\doi}[1]{doi: #1}\else
  \providecommand{\doi}{doi: \begingroup \urlstyle{rm}\Url}\fi

\bibitem[Abbasnejad et~al.(2019)Abbasnejad, Shi, Hengel, and
  Liu]{abbasnejad2019generative}
M~Ehsan Abbasnejad, Qinfeng Shi, Anton van~den Hengel, and Lingqiao Liu.
\newblock A generative adversarial density estimator.
\newblock In \emph{Proceedings of the IEEE/CVF Conference on Computer Vision
  and Pattern Recognition}, pages 10782--10791, 2019.

\bibitem[Abbasnejad et~al.(2020)Abbasnejad, Shi, van~den Hengel, and
  Liu]{abbasnejad2020gade}
M~Ehsan Abbasnejad, Javen Shi, Anton van~den Hengel, and Lingqiao Liu.
\newblock Gade: A generative adversarial approach to density estimation and its
  applications.
\newblock \emph{International Journal of Computer Vision}, 128\penalty0
  (10):\penalty0 2731--2743, 2020.

\bibitem[Ackley et~al.(1985)Ackley, Hinton, and Sejnowski]{hinton1985learning}
David~H. Ackley, Geoffrey~E. Hinton, and Terrence~J. Sejnowski.
\newblock A learning algorithm for boltzmann machines.
\newblock \emph{Cognitive Science}, 9\penalty0 (1):\penalty0 147--169, 1985.

\bibitem[Alemi et~al.(2018)Alemi, Fischer, and Dillon]{alemi2018uncertainty}
Alexander~A Alemi, Ian Fischer, and Joshua~V Dillon.
\newblock Uncertainty in the variational information bottleneck.
\newblock In \emph{Uncertainty in Artificial Intelligence Workshop}, 2018.

\bibitem[Arjovsky et~al.(2017)Arjovsky, Chintala, and
  Bottou]{arjovsky2017wasserstein}
Martin Arjovsky, Soumith Chintala, and L{\'e}on Bottou.
\newblock Wasserstein generative adversarial networks.
\newblock In \emph{International conference on machine learning}, pages
  214--223. PMLR, 2017.

\bibitem[Blei et~al.(2017)Blei, Kucukelbir, and McAuliffe]{blei2017variational}
David~M Blei, Alp Kucukelbir, and Jon~D McAuliffe.
\newblock Variational inference: A review for statisticians.
\newblock \emph{Journal of the American statistical Association}, 112\penalty0
  (518):\penalty0 859--877, 2017.

\bibitem[Burda et~al.(2015)Burda, Grosse, and Salakhutdinov]{pmlr-v38-burda15}
Yuri Burda, Roger Grosse, and Ruslan Salakhutdinov.
\newblock {Accurate and conservative estimates of MRF log-likelihood using
  reverse annealing}.
\newblock In \emph{Proceedings of the Eighteenth International Conference on
  Artificial Intelligence and Statistics}, volume~38, pages 102--110. PMLR,
  09--12 May 2015.

\bibitem[Che et~al.(2020)Che, Zhang, Sohl-Dickstein, Larochelle, Paull, Cao,
  and Bengio]{che2020your}
Tong Che, Ruixiang Zhang, Jascha Sohl-Dickstein, Hugo Larochelle, Liam Paull,
  Yuan Cao, and Yoshua Bengio.
\newblock Your {GAN} is secretly an energy-based model and you should use
  discriminator driven latent sampling.
\newblock In \emph{Advances in Neural Information Processing Systems},
  volume~33, pages 12275--12287, 2020.

\bibitem[Choi et~al.(2018)Choi, Jang, and Alemi]{choi2018waic}
Hyunsun Choi, Eric Jang, and Alexander~A Alemi.
\newblock {WAIC}, but why? {G}enerative ensembles for robust anomaly detection.
\newblock \emph{arXiv preprint arXiv:1810.01392}, 2018.

\bibitem[Choi et~al.(2024)Choi, Choi, and Kang]{choi2024generative}
Jaemoo Choi, Jaewoong Choi, and Myungjoo Kang.
\newblock Generative modeling through the semi-dual formulation of unbalanced
  optimal transport.
\newblock \emph{Advances in Neural Information Processing Systems}, 36, 2024.

\bibitem[Dai et~al.(2017)Dai, Almahairi, Bachman, Hovy, and
  Courville]{dai2017calibrating}
Zihang Dai, Amjad Almahairi, Philip Bachman, Eduard Hovy, and Aaron Courville.
\newblock Calibrating energy-based generative adversarial networks.
\newblock In \emph{International Conference on Learning Representations}, 2017.

\bibitem[Dieng et~al.(2019)Dieng, Ruiz, Blei, and Titsias]{dieng2019prescribed}
Adji~B Dieng, Francisco~JR Ruiz, David~M Blei, and Michalis~K Titsias.
\newblock Prescribed generative adversarial networks.
\newblock \emph{arXiv preprint arXiv:1910.04302}, 2019.

\bibitem[Dinh et~al.(2015)Dinh, Krueger, and Bengio]{dinh2014nice}
Laurent Dinh, David Krueger, and Yoshua Bengio.
\newblock Nice: Non-linear independent components estimation.
\newblock In \emph{International Conference on Learning Representations
  Workshop}, 2015.

\bibitem[Du and Mordatch(2019)]{du2019implicit}
Yilun Du and Igor Mordatch.
\newblock Implicit generation and modeling with energy based models.
\newblock In \emph{Advances in Neural Information Processing Systems},
  volume~32, 2019.

\bibitem[Frellsen et~al.(2016)Frellsen, Winther, Ghahramani, and
  Ferkinghoff-Borg]{pmlr-v51-frellsen16}
Jes Frellsen, Ole Winther, Zoubin Ghahramani, and Jesper Ferkinghoff-Borg.
\newblock Bayesian generalised ensemble markov chain monte carlo.
\newblock In \emph{Proceedings of the 19th International Conference on
  Artificial Intelligence and Statistics}, volume~51, pages 408--416. PMLR,
  09--11 May 2016.

\bibitem[Gao et~al.(2021)Gao, Song, Poole, Wu, and Kingma]{gao2020learning}
Ruiqi Gao, Yang Song, Ben Poole, Ying~Nian Wu, and Diederik~P Kingma.
\newblock Learning energy-based models by diffusion recovery likelihood.
\newblock \emph{International Conference on Learning Representations}, 2021.

\bibitem[Geng et~al.(2021)Geng, Wang, Gao, Frellsen, and
  Hauberg]{geng2021bounds}
Cong Geng, Jia Wang, Zhiyong Gao, Jes Frellsen, and S{\o}ren Hauberg.
\newblock Bounds all around: training energy-based models with bidirectional
  bounds.
\newblock \emph{Advances in Neural Information Processing Systems},
  34:\penalty0 19808--19821, 2021.

\bibitem[Geng et~al.(2024)Geng, Han, Jiang, Zhang, Chen, Hauberg, and
  Li]{geng2024improving}
Cong Geng, Tian Han, Peng-Tao Jiang, Hao Zhang, Jinwei Chen, S{\o}ren Hauberg,
  and Bo~Li.
\newblock Improving adversarial energy-based model via diffusion process.
\newblock \emph{Proceedings of the 41th International Conference on Machine
  Learning}, 2024.

\bibitem[Goodfellow et~al.(2020)Goodfellow, Pouget-Abadie, Mirza, Xu,
  Warde-Farley, Ozair, Courville, and Bengio]{goodfellow2020generative}
Ian Goodfellow, Jean Pouget-Abadie, Mehdi Mirza, Bing Xu, David Warde-Farley,
  Sherjil Ozair, Aaron Courville, and Yoshua Bengio.
\newblock Generative adversarial networks.
\newblock \emph{Communications of the ACM}, 63\penalty0 (11):\penalty0
  139--144, 2020.

\bibitem[Grathwohl et~al.(2021)Grathwohl, Kelly, Hashemi, Norouzi, Swersky, and
  Duvenaud]{nomcmc}
Will~Sussman Grathwohl, Jacob~Jin Kelly, Milad Hashemi, Mohammad Norouzi, Kevin
  Swersky, and David Duvenaud.
\newblock No {MCMC} for me: Amortized sampling for fast and stable training of
  energy-based models.
\newblock In \emph{International Conference on Learning Representations}, 2021.

\bibitem[Grosse et~al.(2013)Grosse, Maddison, and
  Salakhutdinov]{grosse2013annealing}
Roger~B Grosse, Chris~J Maddison, and Russ~R Salakhutdinov.
\newblock Annealing between distributions by averaging moments.
\newblock \emph{Advances in Neural Information Processing Systems}, 26, 2013.

\bibitem[Gulrajani et~al.(2017)Gulrajani, Ahmed, Arjovsky, Dumoulin, and
  Courville]{gulrajani2017improved}
Ishaan Gulrajani, Faruk Ahmed, Martin Arjovsky, Vincent Dumoulin, and Aaron~C
  Courville.
\newblock Improved training of wasserstein gans.
\newblock \emph{Advances in neural information processing systems}, 30, 2017.

\bibitem[Gutmann and Hyv{\"a}rinen(2010)]{gutmann2010noise}
Michael Gutmann and Aapo Hyv{\"a}rinen.
\newblock Noise-contrastive estimation: A new estimation principle for
  unnormalized statistical models.
\newblock In \emph{Proceedings of the Thirteenth International Conference on
  Artificial Intelligence and Statistics}, pages 297--304. JMLR Workshop and
  Conference Proceedings, 2010.

\bibitem[Han et~al.(2019)Han, Nijkamp, Fang, Hill, Zhu, and
  Wu]{han2019divergence}
Tian Han, Erik Nijkamp, Xiaolin Fang, Mitch Hill, Song-Chun Zhu, and Ying~Nian
  Wu.
\newblock Divergence triangle for joint training of generator model,
  energy-based model, and inferential model.
\newblock In \emph{Proceedings of the IEEE/CVF Conference on Computer Vision
  and Pattern Recognition}, pages 8670--8679, 2019.

\bibitem[Han et~al.(2020)Han, Nijkamp, Zhou, Pang, Zhu, and Wu]{han2020joint}
Tian Han, Erik Nijkamp, Linqi Zhou, Bo~Pang, Song-Chun Zhu, and Ying~Nian Wu.
\newblock Joint training of variational auto-encoder and latent energy-based
  model.
\newblock In \emph{Proceedings of the IEEE/CVF Conference on Computer Vision
  and Pattern Recognition}, pages 7978--7987, 2020.

\bibitem[Havtorn et~al.(2021)Havtorn, Frellsen, Hauberg, and
  Maal{\o}e]{havtorn2021hierarchical}
Jakob~D Havtorn, Jes Frellsen, S{\o}ren Hauberg, and Lars Maal{\o}e.
\newblock Hierarchical vaes know what they don’t know.
\newblock In \emph{International Conference on Machine Learning}, pages
  4117--4128. PMLR, 2021.

\bibitem[Hendrycks and Gimpel(2017)]{hendrycks2016baseline}
Dan Hendrycks and Kevin Gimpel.
\newblock A baseline for detecting misclassified and out-of-distribution
  examples in neural networks.
\newblock In \emph{International Conference on Learning Representations}, 2017.

\bibitem[Hendrycks et~al.(2019)Hendrycks, Mazeika, and
  Dietterich]{hendrycks2018deep}
Dan Hendrycks, Mantas Mazeika, and Thomas Dietterich.
\newblock Deep anomaly detection with outlier exposure.
\newblock In \emph{International Conference on Learning Representations}, 2019.

\bibitem[Hinton(2002)]{hinton2002training}
Geoffrey~E Hinton.
\newblock Training products of experts by minimizing contrastive divergence.
\newblock \emph{Neural computation}, 14\penalty0 (8):\penalty0 1771--1800,
  2002.

\bibitem[Hinton and Sejnowski(1983)]{hinton1983optimal}
Geoffrey~E Hinton and Terrence~J Sejnowski.
\newblock Optimal perceptual inference.
\newblock In \emph{Proceedings of the IEEE conference on Computer Vision and
  Pattern Recognition}, volume 448, 1983.

\bibitem[Hinton et~al.(2006)Hinton, Osindero, and Teh]{hinton2006fast}
Geoffrey~E Hinton, Simon Osindero, and Yee-Whye Teh.
\newblock A fast learning algorithm for deep belief nets.
\newblock \emph{Neural computation}, 18\penalty0 (7):\penalty0 1527--1554,
  2006.

\bibitem[Hjelm et~al.(2019)Hjelm, Fedorov, Lavoie-Marchildon, Grewal, Bachman,
  Trischler, and Bengio]{hjelm2018learning}
R~Devon Hjelm, Alex Fedorov, Samuel Lavoie-Marchildon, Karan Grewal, Phil
  Bachman, Adam Trischler, and Yoshua Bengio.
\newblock Learning deep representations by mutual information estimation and
  maximization.
\newblock In \emph{International Conference on Learning Representations}, 2019.

\bibitem[Ho et~al.(2020)Ho, Jain, and Abbeel]{ho2020denoising}
Jonathan Ho, Ajay Jain, and Pieter Abbeel.
\newblock Denoising diffusion probabilistic models.
\newblock In \emph{Advances in Neural Information Processing Systems},
  volume~33, pages 6840--6851, 2020.

\bibitem[Hopfield(1982)]{Hopfield2554}
J~J Hopfield.
\newblock Neural networks and physical systems with emergent collective
  computational abilities.
\newblock \emph{Proceedings of the National Academy of Sciences}, 79\penalty0
  (8):\penalty0 2554--2558, 1982.
\newblock ISSN 0027-8424.

\bibitem[Hutchinson(1989)]{hutchinson1989stochastic}
Michael~F Hutchinson.
\newblock A stochastic estimator of the trace of the influence matrix for
  laplacian smoothing splines.
\newblock \emph{Communications in Statistics-Simulation and Computation},
  18\penalty0 (3):\penalty0 1059--1076, 1989.

\bibitem[Hyv{\"a}rinen(2005)]{hyvarinen2005estimation}
Aapo Hyv{\"a}rinen.
\newblock Estimation of non-normalized statistical models by score matching.
\newblock \emph{Journal of Machine Learning Research}, 6\penalty0 (4), 2005.

\bibitem[Kan et~al.(2022)Kan, L{\"u}, Wang, Zhang, Zhu, Huang, Guo, and
  Snoussi]{kan2022bi}
Ge~Kan, Jinhu L{\"u}, Tian Wang, Baochang Zhang, Aichun Zhu, Lei Huang, Guodong
  Guo, and Hichem Snoussi.
\newblock Bi-level doubly variational learning for energy-based latent variable
  models.
\newblock In \emph{Proceedings of the IEEE/CVF Conference on Computer Vision
  and Pattern Recognition}, pages 18460--18469, 2022.

\bibitem[Kang and Park(2020)]{kang2020ContraGAN}
Minguk Kang and Jaesik Park.
\newblock {ContraGAN}: Contrastive learning for conditional image generation.
\newblock In \emph{Advances in Neural Information Processing Systems},
  volume~33, pages 21357--21369, 2020.

\bibitem[Kim et~al.(2021)Kim, Shin, Song, Kang, and Moon]{kim2021soft}
Dongjun Kim, Seungjae Shin, Kyungwoo Song, Wanmo Kang, and Il-Chul Moon.
\newblock Soft truncation: A universal training technique of score-based
  diffusion model for high precision score estimation.
\newblock In \emph{Proceedings of the 38th International Conference on Machine
  Learning}, 2021.

\bibitem[Kim and Bengio(2016)]{kim2016deep}
Taesup Kim and Yoshua Bengio.
\newblock Deep directed generative models with energy-based probability
  estimation.
\newblock \emph{arXiv preprint arXiv:1606.03439}, 2016.

\bibitem[{}Knyazev(2001)]{knyazev2001toward}
Andrew~V {}Knyazev.
\newblock Toward the optimal preconditioned eigensolver: Locally optimal block
  preconditioned conjugate gradient method.
\newblock \emph{SIAM journal on scientific computing}, 23\penalty0
  (2):\penalty0 517--541, 2001.

\bibitem[Krizhevsky et~al.(2009)Krizhevsky, Hinton,
  et~al.]{krizhevsky2009learning}
Alex Krizhevsky, Geoffrey Hinton, et~al.
\newblock Learning multiple layers of features from tiny images.
\newblock Technical report, University of Toronto, 2009.

\bibitem[Kumar et~al.(2020)Kumar, Poole, and Murphy]{kumar2020regularized}
Abhishek Kumar, Ben Poole, and Kevin Murphy.
\newblock Regularized autoencoders via relaxed injective probability flow.
\newblock In \emph{International Conference on Artificial Intelligence and
  Statistics}, pages 4292--4301. PMLR, 2020.

\bibitem[Kumar et~al.(2019)Kumar, Ozair, Goyal, Courville, and
  Bengio]{kumar2019maximum}
Rithesh Kumar, Sherjil Ozair, Anirudh Goyal, Aaron Courville, and Yoshua
  Bengio.
\newblock Maximum entropy generators for energy-based models.
\newblock \emph{arXiv preprint arXiv:1901.08508}, 2019.

\bibitem[LeCun et~al.(2006)LeCun, Chopra, Hadsell, Ranzato, and
  Huang]{lecun2006tutorial}
Yann LeCun, Sumit Chopra, Raia Hadsell, M~Ranzato, and F~Huang.
\newblock A tutorial on energy-based learning.
\newblock \emph{Predicting structured data}, 1\penalty0 (0), 2006.

\bibitem[Lee et~al.(2023)Lee, Jeong, Park, and Shin]{lee2023guiding}
Hankook Lee, Jongheon Jeong, Sejun Park, and Jinwoo Shin.
\newblock Guiding energy-based models via contrastive latent variables.
\newblock In \emph{International Conference on Learning Representations}, 2023.

\bibitem[Li and Turner(2018)]{li2017gradient}
Yingzhen Li and Richard~E. Turner.
\newblock Gradient estimators for implicit models.
\newblock In \emph{International Conference on Learning Representations}, 2018.

\bibitem[Liu et~al.(2015)Liu, Luo, Wang, and Tang]{liu2015deep}
Ziwei Liu, Ping Luo, Xiaogang Wang, and Xiaoou Tang.
\newblock Deep learning face attributes in the wild.
\newblock In \emph{Proceedings of the IEEE international conference on computer
  vision}, pages 3730--3738, 2015.

\bibitem[Luhman and Luhman(2021)]{luhman2021knowledge}
Eric Luhman and Troy Luhman.
\newblock Knowledge distillation in iterative generative models for improved
  sampling speed.
\newblock \emph{arXiv preprint arXiv:2101.02388}, 2021.

\bibitem[Metropolis et~al.(1953)Metropolis, Rosenbluth, Rosenbluth, Teller, and
  Teller]{metropolis1953equation}
Nicholas Metropolis, Arianna~W Rosenbluth, Marshall~N Rosenbluth, Augusta~H
  Teller, and Edward Teller.
\newblock Equation of state calculations by fast computing machines.
\newblock \emph{The journal of chemical physics}, 21\penalty0 (6):\penalty0
  1087--1092, 1953.

\bibitem[Miyato et~al.(2018)Miyato, Kataoka, Koyama, and
  Yoshida]{miyato2018spectral}
Takeru Miyato, Toshiki Kataoka, Masanori Koyama, and Yuichi Yoshida.
\newblock Spectral normalization for generative adversarial networks.
\newblock In \emph{International Conference on Learning Representations}, 2018.

\bibitem[Neal et~al.(2011)]{neal2011mcmc}
Radford~M Neal et~al.
\newblock {MCMC} using hamiltonian dynamics.
\newblock \emph{Handbook of markov chain monte carlo}, 2\penalty0
  (11):\penalty0 2, 2011.

\bibitem[Nijkamp et~al.(2019)Nijkamp, Hill, Zhu, and Wu]{nijkamp2019learning}
Erik Nijkamp, Mitch Hill, Song-Chun Zhu, and Ying~Nian Wu.
\newblock Learning non-convergent non-persistent short-run mcmc toward
  energy-based model.
\newblock \emph{Advances in Neural Information Processing Systems}, 32, 2019.

\bibitem[Nijkamp et~al.(2020)Nijkamp, Hill, Han, Zhu, and
  Wu]{nijkamp2020anatomy}
Erik Nijkamp, Mitch Hill, Tian Han, Song-Chun Zhu, and Ying~Nian Wu.
\newblock On the anatomy of {MCMC}-based maximum likelihood learning of
  energy-based models.
\newblock In \emph{Proceedings of the AAAI Conference on Artificial
  Intelligence}, volume~34, pages 5272--5280, 2020.

\bibitem[Osogami(2017)]{osogami2017boltzmann}
Takayuki Osogami.
\newblock Boltzmann machines and energy-based models.
\newblock \emph{arXiv preprint arXiv:1708.06008}, 2017.

\bibitem[Paszke et~al.(2017)Paszke, Gross, Chintala, Chanan, Yang, DeVito, Lin,
  Desmaison, Antiga, and Lerer]{paszke2017automatic}
Adam Paszke, Sam Gross, Soumith Chintala, Gregory Chanan, Edward Yang, Zachary
  DeVito, Zeming Lin, Alban Desmaison, Luca Antiga, and Adam Lerer.
\newblock Automatic differentiation in pytorch.
\newblock \emph{NIPS 2017 Workshop Autodiff}, 2017.

\bibitem[Radford et~al.(2016)Radford, Metz, and
  Chintala]{radford2015unsupervised}
Alec Radford, Luke Metz, and Soumith Chintala.
\newblock Unsupervised representation learning with deep convolutional
  generative adversarial networks.
\newblock In \emph{International Conference on Learning Representations}, 2016.

\bibitem[Ren et~al.(2019)Ren, Liu, Fertig, Snoek, Poplin, Depristo, Dillon, and
  Lakshminarayanan]{ren2019likelihood}
Jie Ren, Peter~J. Liu, Emily Fertig, Jasper Snoek, Ryan Poplin, Mark Depristo,
  Joshua Dillon, and Balaji Lakshminarayanan.
\newblock Likelihood ratios for out-of-distribution detection.
\newblock In \emph{Advances in Neural Information Processing Systems},
  volume~32, 2019.

\bibitem[Sajjadi et~al.(2018)Sajjadi, Bachem, Lucic, Bousquet, and
  Gelly]{sajjadi2018assessing}
Mehdi~SM Sajjadi, Olivier Bachem, Mario Lucic, Olivier Bousquet, and Sylvain
  Gelly.
\newblock Assessing generative models via precision and recall.
\newblock \emph{Advances in neural information processing systems}, 31, 2018.

\bibitem[Salakhutdinov and Hinton(2009)]{pmlr-v5-salakhutdinov09a}
Ruslan Salakhutdinov and Geoffrey Hinton.
\newblock Deep boltzmann machines.
\newblock In \emph{Proceedings of the Twelth International Conference on
  Artificial Intelligence and Statistics}, volume~5, pages 448--455. PMLR,
  16--18 Apr 2009.

\bibitem[Salakhutdinov and Murray(2008)]{salakhutdinovicml08a}
Ruslan Salakhutdinov and Iain Murray.
\newblock On the quantitative analysis of deep belief networks.
\newblock In \emph{Proceedings of the 25th International Conference on Machine
  Learning}, pages 872--879, 2008.

\bibitem[Salimans et~al.(2017)Salimans, Karpathy, Chen, and
  Kingma]{Salimans_Karpathy_Chen_Kingma_2017}
Tim Salimans, Andrej Karpathy, Xi~Chen, and DiederikP. Kingma.
\newblock Pixelcnn++: Improving the pixelcnn with discretized logistic mixture
  likelihood and other modifications.
\newblock In \emph{International Conference on Learning Representations}, Jan
  2017.

\bibitem[Scellier(2020)]{scellier2020deep}
Benjamin Scellier.
\newblock \emph{A deep learning theory for neural networks grounded in
  physics}.
\newblock PhD thesis, Université de Montréal, Quebec, Canada, 2020.

\bibitem[Shi et~al.(2018)Shi, Sun, and Zhu]{shi2018spectral}
Jiaxin Shi, Shengyang Sun, and Jun Zhu.
\newblock A spectral approach to gradient estimation for implicit
  distributions.
\newblock In \emph{International Conference on Machine Learning}, pages
  4644--4653. PMLR, 2018.

\bibitem[Smolensky(1986)]{10.5555/104279.104290}
P.~Smolensky.
\newblock \emph{Information Processing in Dynamical Systems: Foundations of
  Harmony Theory}, page 194–281.
\newblock MIT Press, Cambridge, MA, USA, 1986.

\bibitem[Song et~al.(2021{\natexlab{a}})Song, Meng, and
  Ermon]{song2020denoising}
Jiaming Song, Chenlin Meng, and Stefano Ermon.
\newblock International conference on learning representations.
\newblock 2021{\natexlab{a}}.

\bibitem[Song and Ermon(2019)]{song2019generative}
Yang Song and Stefano Ermon.
\newblock Generative modeling by estimating gradients of the data distribution.
\newblock \emph{Advances in Neural Information Processing Systems}, 32, 2019.

\bibitem[Song and Ermon(2020)]{song2020improved}
Yang Song and Stefano Ermon.
\newblock Improved techniques for training score-based generative models.
\newblock \emph{Advances in neural information processing systems},
  33:\penalty0 12438--12448, 2020.

\bibitem[Song et~al.(2021{\natexlab{b}})Song, Durkan, Murray, and
  Ermon]{song2021maximum}
Yang Song, Conor Durkan, Iain Murray, and Stefano Ermon.
\newblock Maximum likelihood training of score-based diffusion models.
\newblock \emph{Advances in Neural Information Processing Systems},
  34:\penalty0 1415--1428, 2021{\natexlab{b}}.

\bibitem[Song et~al.(2021{\natexlab{c}})Song, Sohl-Dickstein, Kingma, Kumar,
  Ermon, and Poole]{song2020score}
Yang Song, Jascha Sohl-Dickstein, Diederik~P Kingma, Abhishek Kumar, Stefano
  Ermon, and Ben Poole.
\newblock Score-based generative modeling through stochastic differential
  equations.
\newblock In \emph{International Conference on Learning Representations},
  2021{\natexlab{c}}.

\bibitem[Song et~al.(2023)Song, Dhariwal, Chen, and
  Sutskever]{song2023consistency}
Yang Song, Prafulla Dhariwal, Mark Chen, and Ilya Sutskever.
\newblock Consistency models.
\newblock \emph{Proceedings of the 40th International Conference on Machine
  Learning}, 2023.

\bibitem[Särkkä and Solin(2019)]{appliedSDE}
Simo Särkkä and Arno Solin.
\newblock \emph{Applied Stochastic Differential Equations}.
\newblock Cambridge University Press, Cambridge, UK, 2019.

\bibitem[Thanh-Tung et~al.(2019)Thanh-Tung, Tran, and
  Venkatesh]{thanh2019improving}
Hoang Thanh-Tung, Truyen Tran, and Svetha Venkatesh.
\newblock Improving generalization and stability of generative adversarial
  networks.
\newblock In \emph{International Conference on Learning Representations}, 2019.

\bibitem[Vincent(2011)]{vincent2011connection}
Pascal Vincent.
\newblock A connection between score matching and denoising autoencoders.
\newblock \emph{Neural computation}, 23\penalty0 (7):\penalty0 1661--1674,
  2011.

\bibitem[Wang and Ponce(2021)]{wang2021geometry}
Binxu Wang and Carlos~R Ponce.
\newblock The geometry of deep generative image models and its applications.
\newblock In \emph{International Conference on Learning Representations}, 2021.

\bibitem[Wu et~al.(2018)Wu, Xie, Lu, and Zhu]{wu2018sparse}
Ying~Nian Wu, Jianwen Xie, Yang Lu, and Song-Chun Zhu.
\newblock Sparse and deep generalizations of the frame model.
\newblock \emph{Annals of Mathematical Sciences and Applications}, 3\penalty0
  (1):\penalty0 211--254, 2018.

\bibitem[Xiao et~al.(2022)Xiao, Kreis, and Vahdat]{xiao2021tackling}
Zhisheng Xiao, Karsten Kreis, and Arash Vahdat.
\newblock Tackling the generative learning trilemma with denoising diffusion
  gans.
\newblock \emph{International Conference on Learning Representations}, 2022.

\bibitem[Xie et~al.(2015)Xie, Hu, Zhu, and Wu]{xie2015learning}
Jianwen Xie, Wenze Hu, Song-Chun Zhu, and Ying~Nian Wu.
\newblock Learning sparse {FRAME} models for natural image patterns.
\newblock \emph{International Journal of Computer Vision}, 114\penalty0
  (2):\penalty0 91--112, 2015.

\bibitem[Xie et~al.(2016)Xie, Lu, Zhu, and Wu]{xie2016inducing}
Jianwen Xie, Yang Lu, Song-Chun Zhu, and Ying~Nian Wu.
\newblock Inducing wavelets into random fields via generative boosting.
\newblock \emph{Applied and Computational Harmonic Analysis}, 41\penalty0
  (1):\penalty0 4--25, 2016.

\bibitem[Xie et~al.(2017)Xie, Zhu, and Nian~Wu]{xie2017synthesizing}
Jianwen Xie, Song-Chun Zhu, and Ying Nian~Wu.
\newblock Synthesizing dynamic patterns by spatial-temporal generative convnet.
\newblock In \emph{Proceedings of the ieee conference on computer vision and
  pattern recognition}, pages 7093--7101, 2017.

\bibitem[Xie et~al.(2018)Xie, Zheng, Gao, Wang, Zhu, and Wu]{xie2018learning}
Jianwen Xie, Zilong Zheng, Ruiqi Gao, Wenguan Wang, Song-Chun Zhu, and
  Ying~Nian Wu.
\newblock Learning descriptor networks for 3d shape synthesis and analysis.
\newblock In \emph{Proceedings of the IEEE conference on computer vision and
  pattern recognition}, pages 8629--8638, 2018.

\bibitem[Xie et~al.(2018a)Xie, Lu, Gao, and Wu]{xie2018cooperative}
Jianwen Xie, Yang Lu, Ruiqi Gao, and Ying~Nian Wu.
\newblock Cooperative learning of energy-based model and latent variable model
  via {MCMC} teaching.
\newblock In \emph{Proceedings of the AAAI Conference on Artificial
  Intelligence}, volume~32, 2018a.

\bibitem[Xie et~al.(2018b)Xie, Lu, Gao, Zhu, and Wu]{xie2018cooperative2}
Jianwen Xie, Yang Lu, Ruiqi Gao, Song-Chun Zhu, and Ying~Nian Wu.
\newblock Cooperative training of descriptor and generator networks.
\newblock \emph{IEEE transactions on pattern analysis and machine
  intelligence}, 42\penalty0 (1):\penalty0 27--45, 2018b.

\bibitem[Xie et~al.(2021{\natexlab{a}})Xie, Zheng, Fang, Zhu, and
  Wu]{xie2021cooperative}
Jianwen Xie, Zilong Zheng, Xiaolin Fang, Song-Chun Zhu, and Ying~Nian Wu.
\newblock Cooperative training of fast thinking initializer and slow thinking
  solver for conditional learning.
\newblock \emph{IEEE Transactions on Pattern Analysis and Machine
  Intelligence}, 2021{\natexlab{a}}.

\bibitem[Xie et~al.(2021{\natexlab{b}})Xie, Zheng, and Li]{xie2021learning}
Jianwen Xie, Zilong Zheng, and Ping Li.
\newblock Learning energy-based model with variational auto-encoder as
  amortized sampler.
\newblock In \emph{The Thirty-Fifth AAAI Conference on Artificial Intelligence
  (AAAI)}, volume~2, 2021{\natexlab{b}}.

\bibitem[Zhai et~al.(2016)Zhai, Cheng, Feris, and Zhang]{zhai2016generative}
Shuangfei Zhai, Yu~Cheng, Rogerio Feris, and Zhongfei Zhang.
\newblock Generative adversarial networks as variational training of energy
  based models.
\newblock \emph{arXiv preprint arXiv:1611.01799}, 2016.

\bibitem[Zhu and Mumford(1998)]{zhu1998grade}
Song~Chun Zhu and David Mumford.
\newblock Grade: {G}ibbs reaction and diffusion equations.
\newblock In \emph{Sixth International Conference on Computer Vision (IEEE Cat.
  No. 98CH36271)}, pages 847--854. IEEE, 1998.

\bibitem[Zhu et~al.(1998)Zhu, Wu, and Mumford]{zhu1998filters}
Song~Chun Zhu, Yingnian Wu, and David Mumford.
\newblock Filters, random fields and maximum entropy ({FRAME}): Towards a
  unified theory for texture modeling.
\newblock \emph{International Journal of Computer Vision}, 27\penalty0
  (2):\penalty0 107--126, 1998.

\bibitem[Zhu et~al.(2024)Zhu, Xie, Wu, and Gao]{zhu2023learning}
Yaxuan Zhu, Jianwen Xie, Yingnian Wu, and Ruiqi Gao.
\newblock Learning energy-based models by cooperative diffusion recovery
  likelihood.
\newblock In \emph{International Conference on Learning Representations}, 2024.

\end{thebibliography}


\end{document}